\documentclass[sigconf, authorversion, nonacm]{acmart}
\DeclareUnicodeCharacter{03B5}{\ensuremath{\varepsilon}}
\AtBeginDocument{%
  }
\usepackage{cleveref}
\usepackage{multirow}
\usepackage{multicol}
\usepackage{subcaption}
\usepackage{pifont}
\usepackage{booktabs} % 用于美化表格
\usepackage{graphicx} % 用于旋转文本
\usepackage{array}    % 用于调整表格布局
\usepackage{graphicx} % for \rotatebox
\usepackage{booktabs} % for professional table lines
\setcopyright{acmlicensed}
\copyrightyear{2025}
\acmYear{2025}
\acmDOI{XXXXXXX.XXXXXXX}
\acmConference[KDD '26]{the 32nd ACM SIGKDD Conference on Knowledge Discovery and Data Mining}{August 9--13,2026}{Jeju, South Korea}

\acmBooktitle{Jeju '26: ACM SIGKDD Conference on Knowledge Discovery and Data Mining,
 August 09--13, 2026, Jeju, South Korea}
\acmISBN{978-1-4503-XXXX-X/2018/06}

\acmSubmissionID{130}

\begin{document}

%%
%% The ''title'' command has an optional parameter,
%% allowing the author to define a ''short title'' to be used in page headers.
\title{TCDiff: Triplex Cascaded Diffusion for High-fidelity Multimodal EHRs Generation with Incomplete Clinical Data}

%%
%% The ''author'' command and its associated commands are used to define
%% ''authornote'' and ''authornotemark'' commands
\author{Yandong Yan}
\orcid{0009-0006-6331-4594}
\affiliation{%
  \institution{School of Computer Science, Peking University}
  \city{Beijing}
  \country{China}
}
\email{ai_yan@stu.pku.edu.cn}

\author{Chenxi Li}
\affiliation{%
  \institution{School of Information and Software Engineering, University of Electronic Science and Technology of China}
  \city{Sichuan}
  \country{China}
}
\email{chenxi.li@std.uestc.edu.cn}

\author{Yu Huang}
\authornote{Corresponding Author}
\affiliation{%
  \institution{National Engineering Research Center for Software Engineering, Peking University}
  \city{Beijing}
  \country{China}}
\email{hy@pku.edu.cn}

\author{Dexuan Xu}
\affiliation{%
  \institution{School of Computer Science, Peking University}
  \city{Beijing}
  \country{China}
}
\email{xudexuan@stu.pku.edu.cn}

\author{Jiaqi Zhu}
\authornotemark[1]
\affiliation{%
  \institution{Institute of Software, Chinese Academy of Science}
  \city{Beijing}
  \country{China}
}
\email{zhujq@ios.ac.cn}

\author{Zhongyan Chai}
\affiliation{%
  \institution{School of Software and Microelectronics, Peking University }
  \city{Beijing}
  \country{China}}
\email{chaizhy@stu.pku.edu.cn}

\author{Huamin Zhang}
\affiliation{%
  \institution{Institute of Basic Theory of Chinese Medicine, China Academy of Chinese Medical Sciences}
  \city{Beijing}
  \country{China}}
\email{hmzhang@icmm.ac.cn}

%%\author{Jia Wu}
%%\affiliation{%
 %% \institution{Department of Information Center, Shenzhen %%Traditional Chinese Medicine Hospital}
%%  \city{Shenzhen}
%%  \country{China}}
%%\email{kokyo\_82@163.com}

\renewcommand{\shortauthors}{Yan et al.}

\begin{abstract}
    %%两个并列用and连接 
    The scarcity of large-scale and high-quality electronic health records (EHRs) remains a major bottleneck in biomedical research, especially as large foundation models become increasingly data-hungry. Synthesizing substantial volumes of de-identified and high-fidelity data from existing datasets has emerged as a promising solution. However, existing methods suffer from a series of limitations: they struggle to model the intrinsic properties of heterogeneous multimodal EHR data (e.g., continuous, discrete, and textual modalities), capture the complex dependencies among them, and robustly handle pervasive data incompleteness. These challenges are particularly acute in Traditional Chinese Medicine (TCM).
    To this end, we propose TCDiff (Triplex Cascaded Diffusion Network), a novel EHR generation framework that cascades three diffusion networks to learn the features of real-world EHR data, formatting a multi-stage generative process: Reference Modalities Diffusion, Cross-Modal Bridging, and Target Modality Diffusion.
    Furthermore, to validate our proposed framework, besides two public datasets, we also construct and introduce TCM-SZ1, a novel multimodal EHR dataset for benchmarking. 
    Experimental results show that TCDiff consistently outperforms state-of-the-art baselines by an average of 10\% in data fidelity under various missing rate, while maintaining competitive privacy guarantees. This highlights the effectiveness, robustness, and generalizability of our approach in real-world healthcare scenarios. 
    % Codes and datasets are available at \url{https://anonymous.4open.science/r/TCDiff-31CE}.
\end{abstract}
\maketitle
\begin{figure}[h]
\centering
\fbox{
  \parbox{0.95\linewidth}{
    \centering
    \textbf{Preprint. Under Review.}
  }
}
\end{figure}
\section{Introduction}
\label{sec:intro}
% % Motivation and Background
    Electronic Health Records (EHRs) provide a rich and multimodal foundation for deep learning approaches in precision medicine. These records are essentially multimodal, integrating structured continuous variables (e.g., lab results), categorical signals (e.g., vital signs), and unstructured text (e.g., clinical notes) \cite{mohsen2022artificial,liu2022machine}. The richness and diversity of these data make EHRs a valuable foundation for learning complex representations of patient health states and predicting individualized outcomes \cite{liu2022multimodal,xu2021mufasa,ben2025multi}.

    Despite their potential, the use of EHRs in machine learning is limited by both privacy constraints and the inherent characteristics of clinical data \cite{fernandez2013security,zhang2022m3care}. Strict regulations on protected health information (PHI) reduce access to large and diverse datasets, while the heterogeneity of medical workflows and selective removal of sensitive modalities often result in missing data across modalities. Together, these issues hinder the development of robust and generalizable models for real-world clinical applications \cite{tertulino2024privacy,bani2020privacy,nowrozy2024privacy}.
    %%可以缩短

    One widely adopted strategy to mitigate privacy risks involves de-identification through data perturbation and randomization, which enables institutions to more easily share EHR data publicly \cite{gkoulalas2014publishing}. Yet, naive applications of such techniques have been repeatedly shown to be vulnerable to re-identification attacks, with classic examples demonstrating how anonymized datasets can be reverse-engineered to re-establish identity links \cite{janmey2018re}. As a promising alternative, growing attention has turned to the generation and dissemination of synthetic EHRs—artificially constructed datasets that mimic the statistical and structural properties of real patient records \cite{wang2024challenges}.
    These synthetic datasets hold the potential to significantly expand the effective volume of usable data while maintaining strong resistance to privacy breaches, thereby offering a viable solution for training large-scale medical models that are heavily reliant on high-quality data \cite{karabacak2023embracing}.
    %%可以缩短

    % While synthetic EHR generation has gained traction as a promising solution to data privacy and accessibility concerns
    However, developing high-fidelity models remains a formidable challenge. Early efforts in this domain often focused on generating structured data in isolation, such as sequences of medical codes (e.g., ICDs), using models like GANs and Autoencoders \cite{choi2017generating,torfi2020corgan,yuan2023ehrdiff,zhao2024ctab}. More recently, the field has progressed towards comprehensive multimodal synthesis, with advanced frameworks like diffusion models aiming to capture a more holistic patient profile \cite{li2023generating,kotelnikov2023tabddpm,he2024flexible,cao2024diffusione}. Despite this significant progress, current state-of-the-art methods still suffer from three major limitations that hinder their applicability in real-world clinical scenarios. 
    \textbf{(1) Inadequate modeling of multimodal data heterogeneity.}
    A fundamental limitation of current methods is their insufficient capacity to model the intrinsic characteristics of individual non-continuous modalities. For structured components, many models resort to workarounds like continuous relaxations to handle non-differentiable categorical data. While enabling 
    unified 
    training, it fundamentally compromises the statistical integrity of the data itself \cite{xu2019modeling, choi2017generating, torfi2020corgan}. The challenge is even greater for unstructured text, as most of the current architectures are simply not equipped to synthesize the long-form and context-rich clinical narratives in real-world EHRs, sidestepping the textual data generation problem.
    \textbf{(2) Limited grasp of cross-modal dependencies.} 
    Existing models often lack robust mechanisms for mutual guidance and interaction of multiple modalities, causing crucial associative information to be lost. For instance, some approaches \cite{li2023generating,he2024flexible} rely on a monolithic latent space for continuous and discrete modalities, which creates an information bottleneck \cite{hazarika2020misa} and risks losing fine-grained, modality-specific details. Others \cite{kotelnikov2023tabddpm} employ separate backbones for different modalities but lack an effective cross-modal interaction mechanism. This failure is particularly pronounced when integrating Traditional Chinese Medicine (TCM) text with structured data, where diagnostic reasoning is deeply encoded in the interplay between textual descriptions and structured observations \cite{chan2015challenges,chen2013filling}.
    \textbf{(3) Insufficient attention on missing modalities.} 
    Real-world EHRs frequently exhibit incomplete records, where entire modalities can be absent due to clinical workflow variations or privacy-driven filtering. Most of the current models assume fully observed input or rely on simplistic imputation strategies, which fail to robustly generate synthetic data with various missingness patterns commonly encountered in practice \cite{song2018enriching,kim2014multiple,li2025fedimpute}. The absence of the textual modality is particularly common, posing a significant challenge for existing models in leveraging complementary data in various forms to estimate the true distribution of real-world EHR data.
    
    To bridge these gaps, we ask:
        \begin{quote}
            \textit{Is it possible to design \textbf{a unified generative framework} that can synthesize complete and high-fidelity multimodal EHR data, even when some modalities in existing data are missing?}
        \end{quote}
% Our Solution and Contributions
    In this paper, we propose the Triplex Cascaded Diffusion Network (TCDiff), a framework that decomposes the intricate task of EHR generation into a multi-stage, coarse-to-fine process. By serially engaging diffusion networks for each modality, our framework fullfills \textit{Reference Modalities Diffusion}, \textit{Cross-Modal Bridging}, and \textit{Target Modality Diffusion} in three stages, which can progressively model the unique intrinsic characteristics of diverse modalities, from structured variables to unstructured clinical narratives. The Cross-Modal Bridging stage, in particular, serves as the core mechanism for capturing the intricate dependencies between these modalities by explicitly learning their interactions. Furthermore, the sequential nature of this cascaded architecture inherently endows TCDiff with robustness against missing data,
    as it can adaptively learn the importance of each modality and alleviate the influence of inevitable noise.
    This integrated, multi-stage strategy enables TCDiff to learn and replicate the complex features of real-world EHRs with high fidelity.

    We conduct comprehensive experiments on three real-world EHR datasets—MIMIC-III, eICU, and a newly collected Traditional Chinese Medicine dataset under varying degrees of modality missing rates. Results show that TCDiff consistently outperforms state-of-the-art methods in both generation fidelity and privacy robustness, demonstrating its superior ability to balance utility and privacy in challenging clinical scenarios.
    
    Our approach makes the following key contributions:
\begin{itemize}
    \item We introduce \textbf{TCDiff}, a diffusion-based generative framework tailored for high-fidelity EHR synthesis trained on incomplete  real-world data, enabling robust generation of heterogeneous clinical data. To the best of our knowledge, this is the first work that supports frequently-missing textual modality in EHR generation.
    \item We propose a novel \textbf{triplex cascaded diffusion architecture} that enables deep cross-modal interaction through a three-stage noise control mechanism: Reference Modalities Diffusion (to establish structural consistency), Cross-Modal Bridging (to align semantic information across modalities), and Target Modality Diffusion (to reconstruct fine-grained modality-specific features). This architecture not only enforces the learning of complecated cross-modal dependencies but also provides a flexible and inherent solution to the missing modality problem.
    \item We demonstrate the superiority of TCDiff through extensive experiments on both public benchmarks and our newly constructed dataset, a large-scale and multimodal TCM benchmark of 60,000 records. Our model consistently outperforms state-of-the-art baselines by an average of \textbf{10\%} in data fidelity, achieves stronger resistance to privacy attacks, 
    showcasing remarkable domain generalization ability across diverse Western and Eastern medicine contexts.
\end{itemize}

\section{Background and Related Works}
\label{sec:related}   
    \subsection{Generative Models for Multimodal EHRs}
        \label{sec:related:generative}
        Early research in synthetic EHR generation primarily focused on structured tabular data. Seminal works employing Generative Adversarial Networks (GANs), such as MedGAN~\cite{choi2017generating} and its variants like CorGAN~\cite{torfi2020corgan}, demonstrated the feasibility of generating realistic records. Following this trend, initial applications of diffusion models, like EHRDiff~\cite{yuan2023ehrdiff}, also concentrated on generating single data types, such as sequences of ICD codes. While foundational, these methods largely treated EHRs as a monolithic block of structured data, sidestepping the practical challenge of multimodal synthesis.

        To create more comprehensive profiles, the field has progressed towards synthesizing heterogeneous multimodal data. Models like TabDDPM~\cite{kotelnikov2023tabddpm} represent a significant step, capable of jointly generating numerical and categorical features. However, the core challenge has now shifted to the interaction mechanism between modalities. Some methods~\cite{li2023generating,he2024flexible} fuse modalities by projecting them into a monolithic latent space. This strategy, however, risks creating an information bottleneck and losing fine-grained modality-specific details~\cite{hazarika2020misa}. Other approaches, including TabDDPM, employ separate processing backbones for different data types but lack a robust mechanism for deep cross-modal interaction, thereby failing to capture their complex dependencies.

        The most significant and persistent gap in EHR generation remains the integration of unstructured text. Clinical narratives, such as physician's notes and discharge summaries, contain indispensable contextual information that defines a patient's journey~\cite{liu2022machine}. Despite this, to the best of our knowledge, no existing generative framework is capable of jointly synthesizing heterogeneous structured EHR data with free-form clinical text. This fundamental omission is a major barrier to generating truly high-fidelity, clinically realistic patient profiles, and highlights the urgent need for a unified model that can bridge this modality gap.
        
        \subsection{Incomplete Multimodal EHR}
        \label{sec:related:incomplete_ehr}
        A significant challenge in leveraging multimodal EHR data for generative modeling stems from its inherent incompleteness. In real-world clinical practice, the assumption of complete data availability across all patient samples is frequently violated due to heterogeneous data collection protocols and socioeconomic factors~\cite{zhang2022m3care}. This phenomenon, often referred to as the missing modality problem, is a critical hurdle for developing robust and reliable clinical models~\cite{tang2020democratizing}.

        When facing incomplete datasets, a common solution is to employ an explicit imputation strategy to fill in missing values before model training. Traditional approaches to this task include statistical techniques such as k-Nearest Neighbors (k-NN)~\cite{batista2003analysis} and Multiple Imputation by Chained Equations (MICE)~\cite{van2011mice}, as well as machine learning methods like the random-forest-based MissForest~\cite{stekhoven2012missforest}.
        However, these conventional imputation strategies demonstrate limited efficacy when applied to complex EHR datasets. Empirical analyses have shown that the accuracy of these methods degrades significantly in the presence of non-random missing patterns or complex cross-modal feature interactions, which are common in healthcare data~\cite{yoon2018estimating,park2019learning,liu2023handling}. The reliance on such potentially flawed pre-processing steps can introduce biases and inaccuracies that propagate through the modeling pipeline. This highlights the critical need for a generative framework that is inherently robust to missing modalities, without depending on a separate and often unreliable imputation phase. \textsc{FLEXGEN} \cite{he2024flexible} handles missing data by aligning features within a unified latent space, is therefore still constrained by the information bottleneck effect, which can lead to the loss of modality-specific details during imputation or generation \cite{hazarika2020misa}. Also, it cannot handle the textual modality for comprehensive understanding and generation of EHR data.

        \begin{figure*}
            \centering
            \includegraphics[width=1\linewidth]{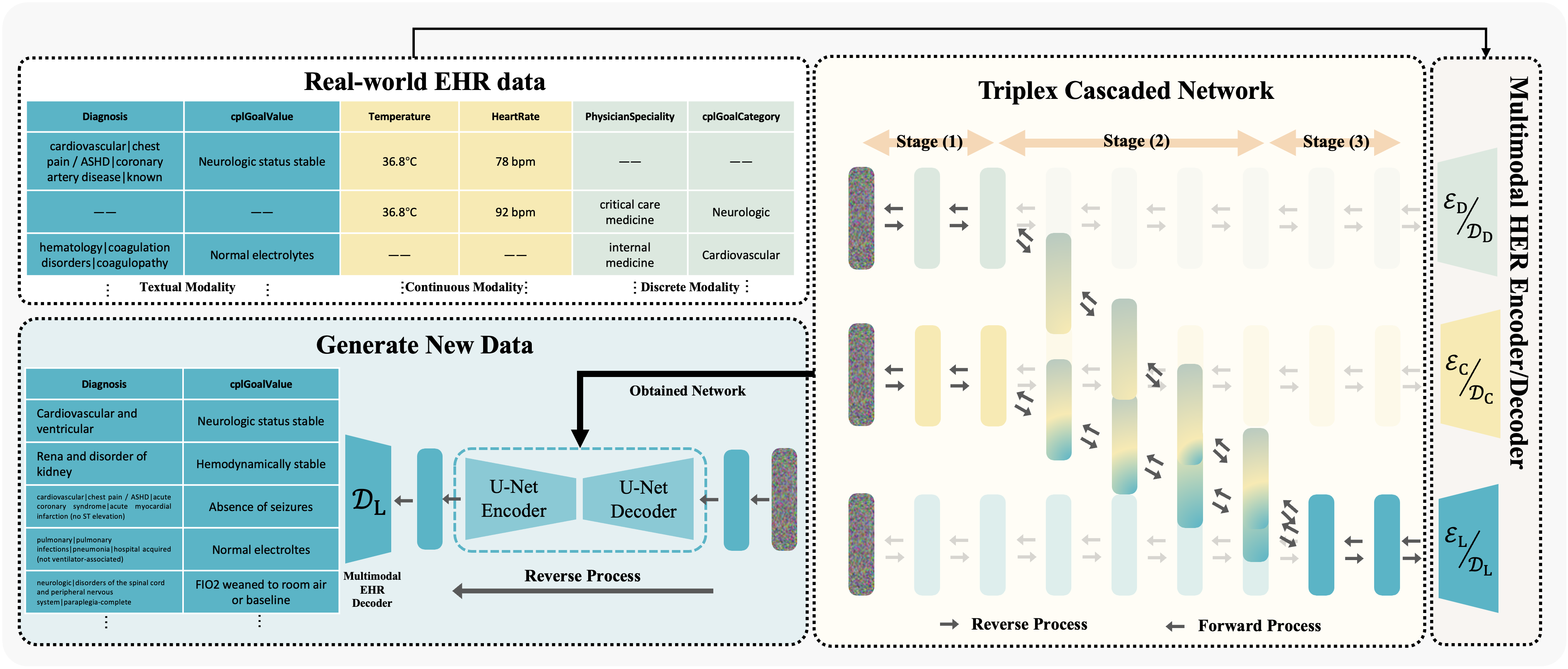}
            \caption{Overview of our proposef TCDiff framework. This figure specifically illustrates the textual modality pathway within TCDiff. The framework comprises three principal components: 
            (1)~a \textit{Multimodal EHR Encoder} that encodes textual, continuous, and discrete modalities into latent representations before forward process; 
            (2)~a \textit{Triplex Cascaded Diffusion Network} with three parallel U-Net-based diffusion branches, each modeling modality-specific noise and supporting cross-modal interaction through a staged generation process; 
            and (3)~a \textit{Multimodal EHR Decoder} that decodes the denoised features after reverse process.
            }
            \label{fig:intro}
        \end{figure*}

\section{Preliminaries}
\label{sec:preliminarise}
    \subsection{Score-based Diffusion Models}
        % Diffusion models are a class of generative models that learn a data distribution by reversing a gradual data corruption process. The core idea is structured around two complementary processes: a fixed forward process that systematically adds noise to data until they becomes pure noise, and a learned reverse process that starts from noise and incrementally denoises the data to generate a clean sample. The key to navigating this reverse process is to estimate the score function \cite{liu2016kernelized,stein1972bound} of the noisy data at each step \cite{song2020score}, which intuitively points in the direction of higher data density. Since the true score is unknown, a score-based model to approximate it, typically by minimizing a score-matching loss function. This allows the model to effectively learn the path from noise back to data.
        % The score function, formally defined as the gradient of the log-probability density, \(s(\mathbf{x})=\nabla_{\mathbf{x}}\log{p(\mathbf{x})}\), mathematically represents the data distribution \(p(\mathbf{x})\) \cite{liu2016kernelized,stein1972bound}. Our goal is to train a score-based model \(s_\theta(\mathbf{x}, t)\) with learnable parameters \(\theta\) to approximate the true score function \(s(\mathbf{x})\) at varying noise levels indexed by time \(t\). This conceptual framework can be formally described using stochastic differential equations (SDEs).

        Score-based diffusion models are a powerful class of generative models that learn a data distribution by first defining a fixed forward process that gradually corrupts data into noise, and then learning a reverse process to transform that noise back into a clean sample. The navigation of this reverse process hinges on the score function, formally defined as the gradient of the log-probability density, \(s(\mathbf{x})=\nabla_{\mathbf{x}}\log{p(\mathbf{x})}\) \cite{liu2016kernelized,stein1972bound}. Intuitively, the score of a noisy data point \(\mathbf{x}(t)\) at time \(t\) points in the direction of higher data density, providing the essential guidance for the denoising at each step. Since the true score function is intractable, we train a neural network, known as a score-based model \(s_\theta(\mathbf{x}, t)\), to approximate it \cite{song2020score}. This is achieved by optimizing a score-matching loss function, which minimizes the discrepancy between the estimated and true scores across all noise levels. This entire generative framework can be rigorously formulated using the mathematics of stochastic differential equations (SDEs), which we detail next.

        \paragraph{\textbf{Perturbing Data with Forward Processing}} Perturbing data with Gaussian noise is essential for generating samples in score-based models. In a diffusion process ${ \mathbf{x}(t) }_{t=0}^{T} \in \mathbb{R}^d$, indexed by $t \in [0, T]$, the probability density $p_t(\mathbf{x})$ is defined. According to \cite{song2020score}, this process follows a stochastic differential equation (SDE):
            \begin{equation}
                \mathrm{d}\mathbf{x} = \mathbf{f}(\mathbf{x}, t) \, \mathrm{d}t + g(t) \, \mathrm{d}\mathbf{w},
                \label{eq:ForwardSDE}
            \end{equation}
        where $\mathbf{f}(\cdot, t)$ denotes the drift coefficient of $\mathbf{x}(t)$, $g(t)$ is a scalar function known as the diffusion coefficient, and $\mathbf{w}$ denotes standard Brownian motion.

        \paragraph{\textbf{Generating Samples by Reverse Processing}} Starting from samples of $\mathbf{x}(T) \sim p_T$ and reversing the \Cref{eq:ForwardSDE} allows us to generate samples $\mathbf{x}(0) \sim p_0$, where $p_0$ represents the true data distribution. As established in the seminal work \cite{anderson1982reverse}, the reverse of a diffusion process is also a diffusion process, obtaining the reverse-time SDE:
            \begin{equation}
            \mathrm{d} \mathbf{x} = \left[ \mathbf{f}(\mathbf{x}, t) - g(t)^2 \nabla_{\mathbf{x}} \log p_t(\mathbf{x}) \right] \mathrm{d}t + g(t) \, \mathrm{d} \bar{\mathbf{w}},
            \label{eq:ReverseSDE}
            \end{equation}
        where $\bar{\mathbf{w}}$ is a standard Wiener process. Once the score of each marginal distribution, $\nabla_{\mathbf{x}} \log p_t(\mathbf{x})$, is known for all $t$, we can derive the reverse diffusion process from \Cref{eq:ReverseSDE} and simulate it to sample from $p_0$.

        \paragraph{\textbf{Score-matching Loss Function}} To optimize the score-matching loss and the score network \( s_\theta \), we leverage the transition kernel \( p_{0t}(\mathbf{x}(t) | \mathbf{x}(0)) \) to approximate the conditional distribution of the state \( \mathbf{x}(t) \) at any time \( t \), given the initial state \( \mathbf{x}(0) \). Specifically, the transition kernel follows a Gaussian distribution, with both its mean and covariance determined by the initial state and the cumulative noise function over time. The equation is as follows:
            \begin{equation}
                p_{0t} (\mathbf{x}(t)|\mathbf{x}(0)) = \mathcal{N} \left(\mathbf{x}(t); \mathbf{x}(0) , \frac{1}{2\log{\sigma}}(1-\varphi(t))\mathbf{I}\right),
                \label{eq:transition-kernel}
            \end{equation}
        where \( \varphi(t) = \sigma^{2t} \) is a decay factor that controls the diffusion process up to time \( t \). Using this, the score-matching loss is formulated as follows:
            \begin{equation}
               \mathcal{L} = \mathbb{E}_{\mathbf{x}, \epsilon \sim \mathcal{N}(\mathbf{0}, \mathbf{I}), t \sim \mathcal{U}(0,T)}  \left\| \frac{\epsilon}{\sqrt{\lambda(t)}} + s_\theta(\mathbf{x}(t), t) \right\|^2_2,
               \label{eq:objective-VP-SDE}
            \end{equation} 
        where \(\epsilon\) is the random noise sampling from \(\mathcal{N}(\mathbf{0},\mathbf{I})\), \( \mathcal{U}(0, T) \) is a uniform distribution over the time interval \( [0, T] \), and \( \lambda(t) = \frac{1}{2\log{\sigma}}(1-\varphi(t)) \) is a weighting function to balance the loss at different time steps.
%----------------------------------------------------------------------------------------
    \subsection{Problem Formulation}
        \label{sec:method:overall}
        Given a real-world tabular EHR dataset \(\mathit{D} = \{(\mathbf{D}_i, \mathbf{C}_i, \mathbf{L}_i)\}_{i=1}^N\) of \(N\) time-dependent electronic health records, \(\mathbf{D}_i \in \mathbb{R}^{S\times d}\) contains discrete features such as categorical or binary variables, \(\mathbf{C}_i \in \mathbb{R}^{S\times c}\) consists of continuous features such as vital signs (e.g., heart rate), and \(\mathbf{L}_i \in \mathbb{R}^{S\times l}\) represents the unstructured textual modality (e.g., clinical notes). Here, \(S\) denotes the sequence length. \(l\), \(c\), and \(d\) denote the feature dimension of the textual, continuous, and discrete modalities, respectively. In practice, not all records contain information from every modality. Specifically, one or two modalities from the discrete (\(\mathbf{D}\)), continuous (\(\mathbf{C}\)), and textual (\(\mathbf{L}\)) may be absent in different records. 
        The goal of this work is to train a generative model on the incomplete dataset \(\mathit{D}\) that learns to approximate the true underlying data distribution. The resulting model can then be used to synthesize an arbitrarily large dataset \( \hat{\mathit{D}} = \{(\hat{\textbf{D}}_i, \hat{\textbf{C}}_i, \hat{\textbf{L}}_i)\}_{i=1}^M \) (\(M \gg N\)) by repeatedly sampling from the learned distribution, with the generated data preserving the statistical properties and the privacy of original records.

\section{The Proposed Method}
    \label{sec:method}  
    In this section, we detail our proposed method, covering the overall framework (\Cref{sec:method:overall}), multimodal EHR encoder (\Cref{sec:method:encoder}), triplex cascaded diffusion network (\Cref{sec:method:network}), multimodal EHR decoder (\Cref{sec:rec}), and the generate process of new data (\Cref{sec:gen}).
%----------------------------------------------------------------------------------------

    %----------------------------------------------------------------------------------------
    \subsection{Overall Framework}
    \label{sec:method:overall}
    Our proposed framework (\Cref{fig:intro}), TCDiff, is architecturally designed to be robust to the missing modalities common in real-world heterogeneous EHRs. It comprises three core components: a Multimodal EHR Encoder, a Triplex Cascaded Diffusion Network, and a Multimodal EHR Decoder. The key to its robustness lies within the Triplex Cascaded Diffusion Network, which moves beyond training three independent diffusion models as in classical score-based models. Instead, our architecture synergistically cascades them to form a new, composite generative process for each modality. To generate a target modality (e.g., discrete data), the reverse diffusion process is initiated within the generative paths of the other available reference modalities (e.g., textual and continuous). As denoising progresses, the process then smoothly transitions to the target modality's own diffusion network for fine-grained refinement. This unique ``cross-network'' trajectory inherently enforces deep cross-modal dependencies, providing a robust foundation for learning from incomplete data.

    \subsection{Multimodal EHR Encoder}
        \label{sec:method:encoder}

       The Multimodal EHR Encoder processes input data from discerete ($\mathbf{D}$), continuous ($\mathbf{C}$) and textual ($\mathbf{L}$) modalities respectively. Each encoder, denoted by $\mathcal{E}_*$ for $* \in \{\mathbf{D}, \mathbf{C}, \mathbf{L}\}$, consists of a 1D convolutional layer to capture local dependencies within each modality. It extracts shallow features from each modality and projects them into latent space of the same dimensionality. For each input dataset $\mathcal{D}$, this framework producs the representation set $\mathcal{X} = \{\mathbf{x}^\mathbf{D}, \mathbf{x}^\mathbf{C}, \mathbf{x}^\mathbf{L}\}$, where $\mathbf{x}^* \in \mathbb{R}^{S \times z }$, for $* \in \{\mathbf{D}, \mathbf{C}, \mathbf{L}\}$, with $S$ denoting the sequence length and  $z$ denoting the latent dimensionality. 
    % Following \cite{imder} to manage missing modalities, we ensure each sample retains at least one modality feature, though the specific available modalities may vary across samples.

    \begin{figure}
        \centering
        \includegraphics[width=1\linewidth]{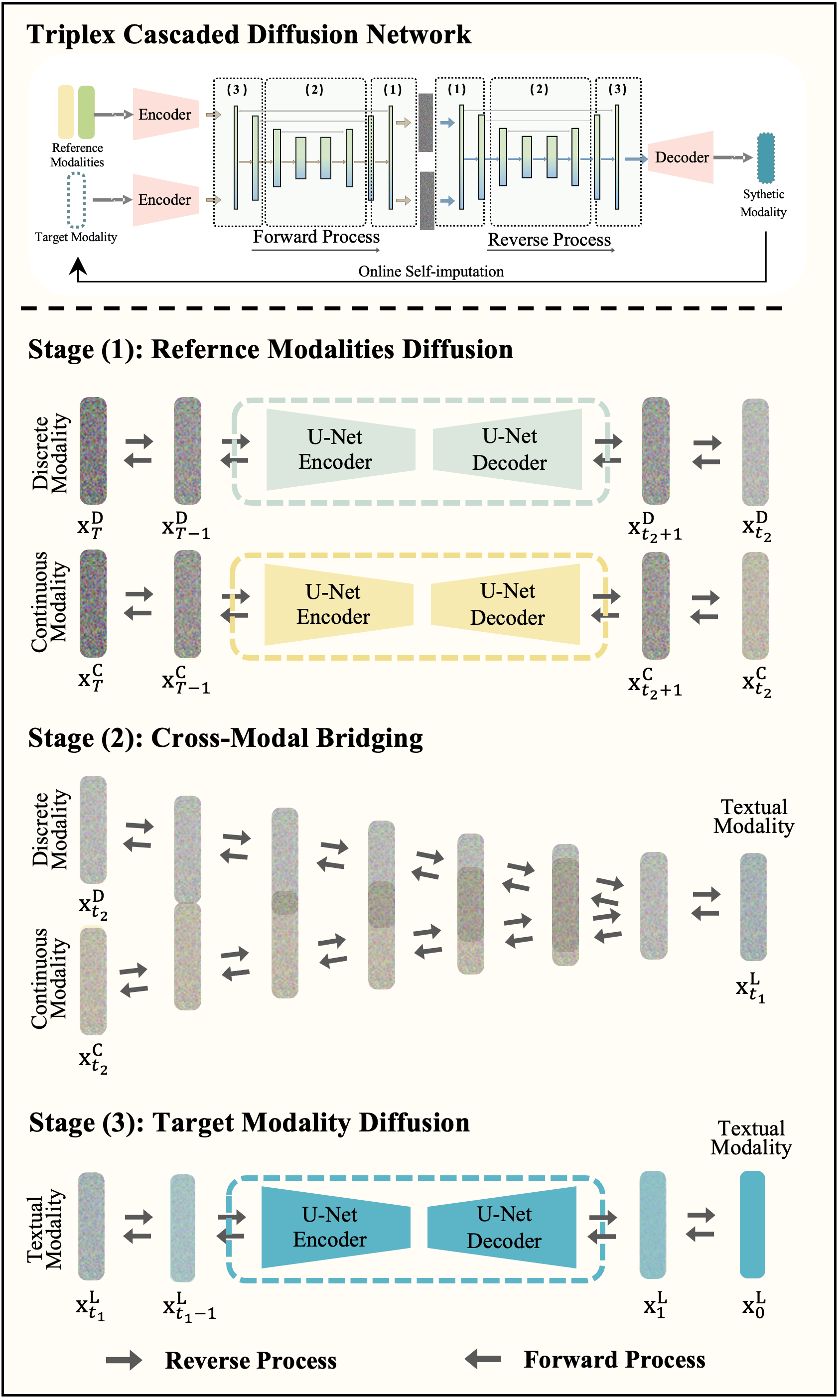}
        \caption{Triplex Cascaded Design of TCDiff.
        We take the textual modality as an example. The architecture consists: 
        (1)~\emph{Reference Modalities Diffusion}, where modality-specific U-Net branches denoise the base discrete and continuous modalities to provide coarse priors; 
        (2)~\emph{Cross-Modal Bridging}, where a cross-modal U-Net fuses information across modalities to enhance representation alignment; 
        and (3)~\emph{Target Modality Diffusion}, where a final U-Net branch reconstructs the target discrete modality with improved fidelity. 
        Each modality is processed in a cascaded manner, allowing the model to progressively generate privacy-preserving synthetic EHRs.
    }
    \label{fig:method}
    \end{figure}

    \subsection{Triplex Cascaded Diffusion Network}
        \label{sec:method:network}  
        To accurately model the complex interplay among different facets of a patient's clinical record, such as continuous lab values, discrete diagnostic codes, and unstructured physician's notes, we introduce the Triplex Cascaded Diffusion Network. Instead of modeling each modality in isolation, our architecture synergistically cascades three modality-specific diffusion processes. The generation of any single modality is initiated and guided by the others before a scheduled transition to its own network for refinement. 
        This cascaded design inherently enforces clinically meaningful dependencies among multimodal data, e.g., ensuring that generated lab results align with the textual diagnostic narrative. This collaborative strategy is the key to generating high-fidelity EHR data and provides innate robustness against missing modalities.

        Specifically, for any target modality \(\mathbf{x}^*\) (where \(^* \in \{\mathbf{D}, \mathbf{C}, \mathbf{L}\}\)), the generation is based on a score-based diffusion model. 
        % The forward process gradually perturbs the data towards noise according to a predefined SDE (as in \Cref{eq:ForwardSDE}). The reverse process, which generates data from noise, is defined by the corresponding reverse-time SDE (as in \Cref{eq:ReverseSDE}). The key to this process is accurately estimating the score function \(\nabla_{\mathbf{x}} \log p_t(\mathbf{x}^*)\) using a neural network \(s_\theta\). 
        Compared to classical forward ande reverse diffusion processes introduced in \Cref{sec:preliminarise}, the innovation of our work lies in the specific architecture of this score network. The goal is to learn the corresponding score function of each modality in its forward process.

        To train this network on real-world clinical data, which is often incomplete, we employ an online self-imputation strategy. For any given patient record, we denote the set of observed modalities as \(\mathcal{M}_{\text{obs}}\) and the set of missing modalities as \(\mathcal{M}_{\text{mis}}\). The process is initialized with a warm-up phase, where the TCDiff model is first trained briefly on data with missing modalities which are simply zero-filled. Subsequently, we begin an iterative process: at the start of each training epoch, the current state of the TCDiff model is used to re-impute all missing modalities \(\forall m \in \mathcal{M}^*_{\text{mis}}\) by generating them conditioned on the observed ones \(\mathcal{M}^*_{\text{obs}}\). The model is then trained for one epoch on this newly completed, higher-fidelity dataset. This iterative refinement allows the model's generative quality and the imputed data's fidelity to mutually enhance each other, enabling robust learning directly from the partial or fragmented records typical of clinical practice.

        \begin{table*}[htbp]
            \centering
            \resizebox{\linewidth}{!}{
            \begin{tabular}{l|l|cccc}
            \toprule
            \textbf{\rotatebox{90}{}} &
              \textbf{Model} &
              \( r_{\text{miss}}= 0\% \) &
              \( r_{\text{miss}}= 10\% \) &
              \( r_{\text{miss}}= 20\% \) &
              \( r_{\text{miss}}= 30\% \) \\ \cmidrule{1-6} 
            \multirow{7}{*}{\rotatebox{90}{MIMIC-III}} &
              MedGAN \cite{choi2017generating} &
              \(0.532_{\pm0.061} / 1.023_{\pm0.072} / 1.428_{\pm0.095}\) &
              \(0.488_{\pm0.063} / 1.102_{\pm0.075} / 1.512_{\pm0.097}\) &
              \(0.444_{\pm0.065} / 1.190_{\pm0.079} / 1.620_{\pm0.101}\) &
              \(0.396_{\pm0.068} / 1.278_{\pm0.082} / 1.720_{\pm0.105}\) \\
             &
              CorGAN \cite{torfi2020corgan} &
              \(0.587_{\pm0.058} / 0.917_{\pm0.068} / 1.206_{\pm0.086}\) &
              \(0.543_{\pm0.059} / 0.957_{\pm0.070} / 1.274_{\pm0.088}\) &
              \(0.499_{\pm0.062} / 1.102_{\pm0.072} / 1.344_{\pm0.091}\) &
              \(0.455_{\pm0.064} / 1.247_{\pm0.075} / 1.420_{\pm0.094}\) \\
             &
              EHR-M-GAN \cite{li2023generating} &
              \(0.613_{\pm0.054} / 0.812_{\pm0.061} / 1.084_{\pm0.073}\) &
              \(0.569_{\pm0.055} / 0.857_{\pm0.063} / 1.124_{\pm0.075}\) &
              \(0.525_{\pm0.057} / 0.902_{\pm0.065} / 1.184_{\pm0.078}\) &
              \(0.481_{\pm0.058} / 0.947_{\pm0.068} / 1.244_{\pm0.081}\) \\
             &
              TabDDPM \cite{kotelnikov2023tabddpm} &
              \(0.648_{\pm0.050} / 0.739_{\pm0.055} / 0.949_{\pm0.067}\) &
              \(0.604_{\pm0.051} / 0.781_{\pm0.056} / 1.028_{\pm0.069}\) &
              \(0.560_{\pm0.053} / 0.826_{\pm0.058} / 1.166_{\pm0.071}\) &
              \(0.516_{\pm0.054} / 0.871_{\pm0.060} / 1.306_{\pm0.074}\) \\
             &
              LDM \cite{rombach2022high} &
              \(0.684_{\pm0.048} / 0.691_{\pm0.048} / 0.802_{\pm0.061}\) &
              \(0.640_{\pm0.049} / 0.745_{\pm0.050} / 0.840_{\pm0.063}\) &
              \(0.596_{\pm0.050} / 0.796_{\pm0.051} / 0.887_{\pm0.065}\) &
              \(0.552_{\pm0.052} / 0.847_{\pm0.053} / 0.924_{\pm0.068}\) \\
             &
              FLEXGEN\_EHR \cite{he2024flexible} &
              \(0.729_{\pm0.045} / 0.668_{\pm0.043} / 0.735_{\pm0.052}\) &
              \(0.715_{\pm0.046} / 0.692_{\pm0.044} / 0.767_{\pm0.053}\) &
              \(0.704_{\pm0.046} / 0.740_{\pm0.045} / 0.819_{\pm0.055}\) &
              \(0.690_{\pm0.048} / 0.788_{\pm0.047} / 0.871_{\pm0.057}\) \\
             &
              \textbf{TCDiff (Ours)} &
              \(\mathbf{0.746_{\pm0.042}} / \mathbf{0.468_{\pm0.041}} / \mathbf{0.636_{\pm0.048}}\) &
              \(\mathbf{0.741_{\pm0.042}} / \mathbf{0.497_{\pm0.042}} / \mathbf{0.664_{\pm0.049}}\) &
              \(\mathbf{0.735_{\pm0.043}} / \mathbf{0.526_{\pm0.043}} / \mathbf{0.692_{\pm0.051}}\) &
              \(\mathbf{0.729_{\pm0.044}} / \mathbf{0.555_{\pm0.044}} / \mathbf{0.720_{\pm0.052}}\) \\ \midrule
            \multirow{7}{*}{\rotatebox{90}{eICU}} &
              MedGAN \cite{choi2017generating} &
              \(0.521_{\pm0.065} / 1.062_{\pm0.078} / 1.462_{\pm0.099}\) &
              \(0.477_{\pm0.067} / 1.113_{\pm0.080} / 1.554_{\pm0.101}\) &
              \(0.433_{\pm0.069} / 1.164_{\pm0.082} / 1.646_{\pm0.104}\) &
              \(0.389_{\pm0.071} / 1.215_{\pm0.085} / 1.738_{\pm0.107}\) \\
             &
              CorGAN \cite{torfi2020corgan} &
              \(0.576_{\pm0.060} / 0.954_{\pm0.072} / 1.238_{\pm0.092}\) &
              \(0.532_{\pm0.062} / 1.105_{\pm0.074} / 1.368_{\pm0.094}\) &
              \(0.488_{\pm0.064} / 1.196_{\pm0.077} / 1.480_{\pm0.097}\) &
              \(0.444_{\pm0.066} / 1.307_{\pm0.080} / 1.598_{\pm0.100}\) \\
             &
              EHR-M-GAN \cite{li2023generating} &
              \(0.601_{\pm0.057} / 0.845_{\pm0.066} / 1.117_{\pm0.081}\) &
              \(0.557_{\pm0.058} / 0.894_{\pm0.068} / 1.181_{\pm0.083}\) &
              \(0.513_{\pm0.060} / 0.943_{\pm0.071} / 1.285_{\pm0.086}\) &
              \(0.469_{\pm0.062} / 0.992_{\pm0.073} / 1.369_{\pm0.089}\) \\
             &
              TabDDPM \cite{kotelnikov2023tabddpm} &
              \(0.637_{\pm0.053} / 0.782_{\pm0.062} / 0.978_{\pm0.078}\) &
              \(0.593_{\pm0.054} / 0.824_{\pm0.063} / 1.010_{\pm0.080}\) &
              \(0.549_{\pm0.056} / 0.866_{\pm0.065} / 1.142_{\pm0.082}\) &
              \(0.505_{\pm0.058} / 0.908_{\pm0.067} / 1.274_{\pm0.085}\) \\
             &
              LDM \cite{rombach2022high} &
              \(0.672_{\pm0.051} / 0.721_{\pm0.051} / 0.841_{\pm0.061}\) &
              \(0.628_{\pm0.052} / 0.763_{\pm0.053} / 0.985_{\pm0.064}\) &
              \(0.584_{\pm0.053} / 0.905_{\pm0.056} / 1.129_{\pm0.067}\) &
              \(0.540_{\pm0.055} / 1.047_{\pm0.058} / 1.273_{\pm0.070}\) \\
             &
              FLEXGEN\_EHR \cite{he2024flexible} &
              \(0.713_{\pm0.048} / 0.684_{\pm0.046} / 0.769_{\pm0.055}\) &
              \(0.701_{\pm0.048} / 0.717_{\pm0.047} / 0.811_{\pm0.056}\) &
              \(0.693_{\pm0.049} / 0.745_{\pm0.048} / 0.853_{\pm0.058}\) &
              \(0.681_{\pm0.050} / 0.793_{\pm0.049} / 0.915_{\pm0.060}\) \\
             &
              \textbf{TCDiff (Ours)} &
              {\(\mathbf{0.739_{\pm0.045}} / \mathbf{0.488_{\pm0.044}} / \mathbf{0.648_{\pm0.053}}\)} &
              {\(\mathbf{0.734_{\pm0.045}} / \mathbf{0.517_{\pm0.045}} / \mathbf{0.676_{\pm0.054}}\)} &
              {\(\mathbf{0.729_{\pm0.046}} / \mathbf{0.546_{\pm0.046}} / \mathbf{0.704_{\pm0.055}}\)} &
              {\(\mathbf{0.722_{\pm0.047}} / \mathbf{0.575_{\pm0.047}} / \mathbf{0.732_{\pm0.056}}\)} \\ \midrule
            \multirow{7}{*}{\rotatebox{90}{TCM-SZ1}} &
              MedGAN \cite{choi2017generating} &
              \(0.508_{\pm0.065} / 1.137_{\pm0.078} / 1.493_{\pm0.099}\) &
              \(0.464_{\pm0.068} / 1.288_{\pm0.081} / 1.745_{\pm0.102}\) &
              \(0.420_{\pm0.070} / 1.439_{\pm0.084} / 1.947_{\pm0.105}\) &
              \(0.376_{\pm0.072} / 1.590_{\pm0.087} / 2.149_{\pm0.109}\) \\
             &
              CorGAN \cite{torfi2020corgan} &
              \(0.563_{\pm0.060} / 1.028_{\pm0.072} / 1.284_{\pm0.092}\) &
              \(0.519_{\pm0.062} / 1.179_{\pm0.075} / 1.526_{\pm0.095}\) &
              \(0.475_{\pm0.064} / 1.330_{\pm0.078} / 1.768_{\pm0.099}\) &
              \(0.431_{\pm0.067} / 1.481_{\pm0.081} / 2.010_{\pm0.102}\) \\
             &
              EHR-M-GAN \cite{li2023generating} &
              \(0.613_{\pm0.057} / 0.926_{\pm0.066} / 1.164_{\pm0.081}\) &
              \(0.569_{\pm0.059} / 1.073_{\pm0.069} / 1.406_{\pm0.084}\) &
              \(0.525_{\pm0.061} / 1.220_{\pm0.072} / 1.648_{\pm0.088}\) &
              \(0.481_{\pm0.063} / 1.364_{\pm0.075} / 1.880_{\pm0.091}\) \\
             &
              TabDDPM \cite{kotelnikov2023tabddpm} &
              \(0.639_{\pm0.053} / 0.824_{\pm0.062} / 1.019_{\pm0.078}\) &
              \(0.595_{\pm0.055} / 0.966_{\pm0.065} / 1.257_{\pm0.081}\) &
              \(0.551_{\pm0.057} / 1.108_{\pm0.068} / 1.495_{\pm0.085}\) &
              \(0.507_{\pm0.059} / 1.250_{\pm0.071} / 1.733_{\pm0.089}\) \\
             &
              LDM \cite{rombach2022high} &
              \(0.677_{\pm0.051} / 0.739_{\pm0.051} / 0.901_{\pm0.061}\) &
              \(0.633_{\pm0.052} / 0.881_{\pm0.054} / 1.143_{\pm0.065}\) &
              \(0.589_{\pm0.054} / 1.023_{\pm0.057} / 1.385_{\pm0.069}\) &
              \(0.545_{\pm0.056} / 1.165_{\pm0.060} / 1.627_{\pm0.073}\) \\
             &
              FLEXGEN\_EHR \cite{he2024flexible} &
              \(0.711_{\pm0.048} / 0.682_{\pm0.046} / 0.833_{\pm0.055}\) &
              \(0.702_{\pm0.049} / 0.774_{\pm0.048} / 1.077_{\pm0.058}\) &
              \(0.694_{\pm0.050} / 0.866_{\pm0.050} / 1.317_{\pm0.061}\) &
              \(0.683_{\pm0.051} / 0.958_{\pm0.052} / 1.557_{\pm0.064}\) \\
             &
              \textbf{TCDiff (Ours)} &
              {\(\mathbf{0.751_{\pm0.045}} / \mathbf{0.464_{\pm0.044}} / \mathbf{0.616_{\pm0.053}}\)} &
              {\(\mathbf{0.744_{\pm0.045}} / \mathbf{0.488_{\pm0.045}} / \mathbf{0.644_{\pm0.054}}\)} &
              {\(\mathbf{0.739_{\pm0.046}} / \mathbf{0.521_{\pm0.046}} / \mathbf{0.672_{\pm0.056}}\)} &
              {\(\mathbf{0.732_{\pm0.047}} / \mathbf{0.549_{\pm0.047}} / \mathbf{0.700_{\pm0.058}}\)} \\ \bottomrule
            \end{tabular}
            }
            \caption{Comparison on data fidelity with the state-of-the-arts models on three datasets under high missing rates (0\%-30\%). The values in each cell denote $\mathbf{R^2}$ $(\uparrow)$ / MMD $(\downarrow)$ / MSE $(\downarrow)$. \textbf{Bold} is the best.}
            \label{exp:missing_1}
        \end{table*}
        
        \paragraph{\textbf{Forward Process}}  
        We illustrate the forward process of textual modality as a representative example. We consider a score network \(s^{\textbf{L}}_\theta\) to model the distribution of target modality \(\mathbf{x}^\textbf{L}\) by perturbing it with a common SDE: \(\mathrm{d}\mathbf{x}=\sigma^t \mathrm{d}\mathbf{w}, t\in [0,1]\), and the corresponding reverse-time SDE can be derived from \Cref{eq:ForwardSDE} and \Cref{eq:ReverseSDE} as:
            \begin{equation}
                \mathrm{d}\mathbf{x}^\textbf{L}=-\sigma^{2t}s^\textbf{L}_\theta(\mathbf{x}^\textbf{L}_t,t)\mathrm{d}t + \sigma^t\mathrm{d}\bar{\mathbf{w}}.
                \label{eq:Re}
            \end{equation}
        When the score network is well-trained based on \Cref{eq:objective-VP-SDE}, we can generate target modality via \Cref{eq:Re}. Below, we use a general-purpose Euler-Maruyama numerical method that is based on a simple discretization to the SDE \cite{song2020score}.
    
        As shown in \Cref{fig:method}, the Triplex Diffusion Networks progressively construct cross-modal dependencies (discrete (\( \mathbf{x}^\textbf{D} \)), continuous (\( \mathbf{x}^\textbf{C} \)) and textual (\( \mathbf{x}^\textbf{L} \))) via three stages. Importantly, these diffusion paths are not isolated; they are dynamically intertwined through the cross-modal bridging stage, which facilitates information transfer across modalities during the forward process. We denote the intermediate states of the diffusion process at timestep \( t \) as \( \tilde{\mathbf{x}}_t \). For clarity, we divide the overall forward process into three sequential time stages. This is opposite to the generation process, so from target modality to base modalities.
        % 核对一下 base modality 的说法， 改一下
    
        In the \textit{Target Modality Diffusion Stage} \(t \in (0, t_1]\),  we inject noise into the target reconstruction textual modality (\( \mathbf{x}^{\textbf{L}} \)).  The forward  process \( \tilde{\mathbf{x}}_t \) in this stage is as:
            \begin{equation}
                \tilde{\mathbf{x}}^{\textbf{L}}_t=\mathbf{x}^\textbf{L}_t=\mathbf{x}^\textbf{L}_{t-1}+\sqrt{\sigma^2_{t}-\sigma^2_{t-1}}\cdot\epsilon_{t-1}.
                \label{eq:Forward-region1}
            \end{equation}

        In the \textit{Cross-Modal Bridging Stage}, defined as \( t \in (t_1, t_2] \), the latent representations of selected reconstruct modality \( \mathbf{x}^\textbf{L}_t \) and base modalities \( \mathbf{x}^{\textbf{D}}_t \), \(\mathbf{x}^{\textbf{C}}_t\) undergo a cross-time-step interaction and information transfer process. This stage continues until the terminal point \( t_2 \), enabling dynamic fusion of modality-specific signals over time. The forward trajectory of \( \tilde{\mathbf{x}}^m_t \) within this interval can be formulated as:
            \begin{equation}
                \tilde{\mathbf{x}}^{\textbf{L}}_t = {\mathbf{x}}_{t-1}^\textbf{L} + \sqrt{\sigma^2_{t} - \sigma^2_{t-1}} \, \epsilon_{t-1} - \Phi \left( \mathbf{x}^\textbf{L}_0 - \left(\alpha^\textbf{L}_t \mathbf{x}^{\textbf{D}}_0 + (1-\alpha^\textbf{L}_t) \mathbf{x}^{\textbf{C}}_0 \right) \right),
                \label{eq:Forward-region2}
            \end{equation}
        where \( \Phi \) is a scaling coefficient that modulates the influence of cross-modal priors, whose value is determined by the SDE in \Cref{eq:ForwardSDE}, and is derived as:
        \[
        \Phi = \frac{t - t_1}{t_2 - t_1},
        \]
        \(\alpha^\textbf{L}_t\) is a learnable fusion weight that adaptively balances the contribution of the discrete and continuous modalities. 
        % This fusion mechanism is applied symmetrically across our framework: when generating any target modality, a corresponding learnable parameter dynamically weights the influence of the other two conditioning modalities.
        
        \begin{table*}[htbp]
            \centering
            \resizebox{\linewidth}{!}{
            \begin{tabular}{l|l|cccc}
            \toprule
            \textbf{\rotatebox{90}{}} &
              \textbf{Model} &
              \( r_{\text{miss}}= 40\% \) &
              \( r_{\text{miss}}= 50\% \) &
              \( r_{\text{miss}}= 60\% \) &
              \( r_{\text{miss}}= 67\% \) \\ \cmidrule{1-6} 
            \multirow{7}{*}{\rotatebox{90}{MIMIC-III}} &
              MedGAN \cite{choi2017generating} &
              \(0.352_{\pm0.071} / 1.366_{\pm0.083} / 1.828_{\pm0.106}\) &
              \(0.308_{\pm0.074} / 1.454_{\pm0.087} / 1.944_{\pm0.110}\) &
              \(0.264_{\pm0.078} / 1.542_{\pm0.091} / 2.068_{\pm0.115}\) &
              \(0.231_{\pm0.081} / 1.601_{\pm0.094} / 2.155_{\pm0.119}\) \\
             &
              CorGAN \cite{torfi2020corgan} &
              \(0.411_{\pm0.065} / 1.392_{\pm0.076} / 1.504_{\pm0.095}\) &
              \(0.367_{\pm0.068} / 1.537_{\pm0.080} / 1.596_{\pm0.099}\) &
              \(0.323_{\pm0.071} / 1.682_{\pm0.084} / 1.698_{\pm0.104}\) &
              \(0.291_{\pm0.073} / 1.785_{\pm0.088} / 1.773_{\pm0.108}\) \\
             &
              EHR-M-GAN \cite{li2023generating} &
              \(0.437_{\pm0.059} / 0.992_{\pm0.069} / 1.304_{\pm0.082}\) &
              \(0.393_{\pm0.062} / 1.037_{\pm0.072} / 1.370_{\pm0.085}\) &
              \(0.349_{\pm0.065} / 1.082_{\pm0.075} / 1.442_{\pm0.089}\) &
              \(0.316_{\pm0.067} / 1.115_{\pm0.078} / 1.496_{\pm0.092}\) \\
             &
              TabDDPM \cite{kotelnikov2023tabddpm} &
              \(0.472_{\pm0.055} / 0.916_{\pm0.061} / 1.454_{\pm0.075}\) &
              \(0.428_{\pm0.057} / 0.961_{\pm0.063} / 1.610_{\pm0.078}\) &
              \(0.384_{\pm0.059} / 1.006_{\pm0.066} / 1.774_{\pm0.081}\) &
              \(0.351_{\pm0.061} / 1.039_{\pm0.068} / 1.891_{\pm0.084}\) \\
             &
              LDM \cite{rombach2022high} &
              \(0.508_{\pm0.053} / 0.898_{\pm0.054} / 0.961_{\pm0.069}\) &
              \(0.464_{\pm0.055} / 0.949_{\pm0.056} / 1.001_{\pm0.072}\) &
              \(0.420_{\pm0.057} / 1.000_{\pm0.058} / 1.045_{\pm0.075}\) &
              \(0.387_{\pm0.059} / 1.036_{\pm0.060} / 1.078_{\pm0.078}\) \\
             &
              FLEXGEN\_EHR \cite{he2024flexible} &
              \(0.677_{\pm0.048} / 0.836_{\pm0.048} / 0.923_{\pm0.058}\) &
              \(0.668_{\pm0.049} / 0.884_{\pm0.049} / 0.978_{\pm0.060}\) &
              \(0.654_{\pm0.050} / 0.932_{\pm0.051} / 1.036_{\pm0.062}\) &
              \(0.639_{\pm0.051} / 0.968_{\pm0.052} / 1.079_{\pm0.064}\) \\
             &
              \textbf{TCDiff (Ours)} &
              \(\mathbf{0.718_{\pm0.044}} / \mathbf{0.584_{\pm0.045}} / \mathbf{0.748_{\pm0.052}}\) &
              \(\mathbf{0.712_{\pm0.045}} / \mathbf{0.613_{\pm0.046}} / \mathbf{0.776_{\pm0.054}}\) &
              \(\mathbf{0.707_{\pm0.046}} / \mathbf{0.642_{\pm0.047}} / \mathbf{0.804_{\pm0.055}}\) &
              \(\mathbf{0.695_{\pm0.047}} / \mathbf{0.664_{\pm0.048}} / \mathbf{0.825_{\pm0.056}}\) \\ \midrule
            \multirow{7}{*}{\rotatebox{90}{eICU}} &
              MedGAN \cite{choi2017generating} &
              \(0.345_{\pm0.073} / 1.266_{\pm0.088} / 1.830_{\pm0.110}\) &
              \(0.301_{\pm0.076} / 1.317_{\pm0.091} / 1.922_{\pm0.114}\) &
              \(0.257_{\pm0.079} / 1.368_{\pm0.094} / 2.014_{\pm0.118}\) &
              \(0.224_{\pm0.082} / 1.405_{\pm0.097} / 2.081_{\pm0.122}\) \\
             &
              CorGAN \cite{torfi2020corgan} &
              \(0.400_{\pm0.068} / 1.418_{\pm0.083} / 1.716_{\pm0.103}\) &
              \(0.356_{\pm0.071} / 1.529_{\pm0.087} / 1.834_{\pm0.107}\) &
              \(0.312_{\pm0.074} / 1.640_{\pm0.091} / 1.952_{\pm0.112}\) &
              \(0.279_{\pm0.077} / 1.721_{\pm0.095} / 2.041_{\pm0.116}\) \\
             &
              EHR-M-GAN \cite{li2023generating} &
              \(0.425_{\pm0.064} / 1.041_{\pm0.076} / 1.453_{\pm0.092}\) &
              \(0.381_{\pm0.067} / 1.090_{\pm0.079} / 1.537_{\pm0.096}\) &
              \(0.337_{\pm0.069} / 1.139_{\pm0.082} / 1.621_{\pm0.100}\) &
              \(0.304_{\pm0.072} / 1.176_{\pm0.085} / 1.685_{\pm0.103}\) \\
             &
              TabDDPM \cite{kotelnikov2023tabddpm} &
              \(0.461_{\pm0.060} / 0.950_{\pm0.069} / 1.406_{\pm0.088}\) &
              \(0.417_{\pm0.062} / 0.992_{\pm0.072} / 1.538_{\pm0.091}\) &
              \(0.373_{\pm0.065} / 1.034_{\pm0.075} / 1.670_{\pm0.095}\) &
              \(0.340_{\pm0.067} / 1.065_{\pm0.077} / 1.769_{\pm0.098}\) \\
             &
              LDM \cite{rombach2022high} &
              \(0.496_{\pm0.057} / 1.189_{\pm0.061} / 1.417_{\pm0.073}\) &
              \(0.452_{\pm0.059} / 1.331_{\pm0.064} / 1.561_{\pm0.077}\) &
              \(0.408_{\pm0.062} / 1.473_{\pm0.068} / 1.705_{\pm0.081}\) &
              \(0.375_{\pm0.064} / 1.584_{\pm0.071} / 1.814_{\pm0.084}\) \\
             &
              FLEXGEN\_EHR \cite{he2024flexible} &
              \(0.669_{\pm0.051} / 0.841_{\pm0.050} / 0.977_{\pm0.061}\) &
              \(0.657_{\pm0.052} / 0.889_{\pm0.052} / 1.039_{\pm0.063}\) &
              \(0.645_{\pm0.053} / 0.937_{\pm0.053} / 1.101_{\pm0.065}\) &
              \(0.636_{\pm0.054} / 0.973_{\pm0.055} / 1.148_{\pm0.067}\) \\
             &
              \textbf{TCDiff (Ours)} &
              {\(\mathbf{0.712_{\pm0.048}} / \mathbf{0.604_{\pm0.048}} / \mathbf{0.760_{\pm0.057}}\)} &
              {\(\mathbf{0.706_{\pm0.049}} / \mathbf{0.633_{\pm0.050}} / \mathbf{0.788_{\pm0.059}}\)} &
              {\(\mathbf{0.701_{\pm0.050}} / \mathbf{0.662_{\pm0.051}} / \mathbf{0.816_{\pm0.060}}\)} &
              {\(\mathbf{0.690_{\pm0.051}} / \mathbf{0.684_{\pm0.052}} / \mathbf{0.837_{\pm0.061}}\)} \\ \midrule
            \multirow{7}{*}{\rotatebox{90}{TCM-SZ1}} &
              MedGAN \cite{choi2017generating} &
              \(0.322_{\pm0.075} / 1.751_{\pm0.090} / 2.361_{\pm0.112}\) &
              \(0.278_{\pm0.078} / 1.902_{\pm0.094} / 2.563_{\pm0.116}\) &
              \(0.234_{\pm0.081} / 2.053_{\pm0.098} / 2.765_{\pm0.121}\) &
              \(0.201_{\pm0.084} / 2.164_{\pm0.101} / 2.906_{\pm0.125}\) \\
             &
              CorGAN \cite{torfi2020corgan} &
              \(0.377_{\pm0.069} / 1.642_{\pm0.084} / 2.262_{\pm0.105}\) &
              \(0.333_{\pm0.072} / 1.793_{\pm0.088} / 2.504_{\pm0.110}\) &
              \(0.289_{\pm0.075} / 1.944_{\pm0.092} / 2.746_{\pm0.115}\) &
              \(0.256_{\pm0.078} / 2.055_{\pm0.096} / 2.927_{\pm0.119}\) \\
             &
              EHR-M-GAN \cite{li2023generating} &
              \(0.427_{\pm0.065} / 1.521_{\pm0.078} / 2.132_{\pm0.094}\) &
              \(0.383_{\pm0.068} / 1.668_{\pm0.082} / 2.374_{\pm0.099}\) &
              \(0.339_{\pm0.071} / 1.815_{\pm0.086} / 2.616_{\pm0.104}\) &
              \(0.306_{\pm0.074} / 1.929_{\pm0.089} / 2.801_{\pm0.108}\) \\
             &
              TabDDPM \cite{kotelnikov2023tabddpm} &
              \(0.453_{\pm0.061} / 1.402_{\pm0.074} / 1.981_{\pm0.092}\) &
              \(0.409_{\pm0.064} / 1.544_{\pm0.078} / 2.219_{\pm0.097}\) &
              \(0.365_{\pm0.067} / 1.686_{\pm0.082} / 2.457_{\pm0.102}\) &
              \(0.332_{\pm0.070} / 1.798_{\pm0.085} / 2.634_{\pm0.106}\) \\
             &
              LDM \cite{rombach2022high} &
              \(0.491_{\pm0.058} / 1.317_{\pm0.063} / 1.879_{\pm0.076}\) &
              \(0.447_{\pm0.061} / 1.459_{\pm0.067} / 2.121_{\pm0.081}\) &
              \(0.403_{\pm0.064} / 1.601_{\pm0.071} / 2.363_{\pm0.086}\) &
              \(0.370_{\pm0.066} / 1.712_{\pm0.074} / 2.544_{\pm0.090}\) \\
             &
              FLEXGEN\_EHR \cite{he2024flexible} &
              \(0.662_{\pm0.052} / 1.060_{\pm0.054} / 1.807_{\pm0.067}\) &
              \(0.651_{\pm0.053} / 1.152_{\pm0.056} / 2.047_{\pm0.070}\) &
              \(0.640_{\pm0.055} / 1.244_{\pm0.058} / 2.287_{\pm0.073}\) &
              \(0.632_{\pm0.056} / 1.316_{\pm0.060} / 2.467_{\pm0.076}\) \\
             &
              \textbf{TCDiff (Ours)} &
              {\(\mathbf{0.724_{\pm0.048}} / \mathbf{0.579_{\pm0.048}} / \mathbf{0.728_{\pm0.059}}\)} &
              {\(\mathbf{0.717_{\pm0.049}} / \mathbf{0.607_{\pm0.050}} / \mathbf{0.756_{\pm0.061}}\)} &
              {\(\mathbf{0.709_{\pm0.050}} / \mathbf{0.639_{\pm0.051}} / \mathbf{0.784_{\pm0.062}}\)} &
              {\(\mathbf{0.702_{\pm0.051}} / \mathbf{0.657_{\pm0.052}} / \mathbf{0.805_{\pm0.064}}\)} \\ \bottomrule
            \end{tabular}
            }
            \caption{Comparison on data fidelity with the state-of-the-arts models on three datasets under high missing rates (40\%-67\%). The values in each cell denote $\mathbf{R^2}$ $(\uparrow)$ / MMD $(\downarrow)$ / MSE $(\downarrow)$. \textbf{Bold} is the best.}
            \label{exp:missing_2}
        \end{table*}

        In the \textit{Reference Modalities Diffusion Stage}, corresponding to the diffusion interval \( t \in (t_2, T] \), the injected noise progressively dominates the latent representations of the conditioning modalities, namely \( \mathbf{x}^{\textbf{D}}_t \) and \( \mathbf{x}^{\textbf{C}}_t \), ultimately driving them toward standard Gaussian distributions. The forward latent state \( \tilde{\mathbf{x}}_t \) during this phase is computed as:
            \begin{equation}
                \tilde{\mathbf{x}}^{\textbf{L}}_t = \alpha^\textbf{L}_t \mathbf{x}^{\textbf{D}}_{t-1} + (1-\alpha^\textbf{L}_t) \mathbf{x}^{\textbf{C}}_{t-1} + \sqrt{\sigma^2_{t} - \sigma^2_{t-1}}\, \epsilon_{t-1},
                \label{eq:Forward-region3}
            \end{equation}

        \paragraph{\textbf{Reverse Process}} The generation process in TCDiff is realized by numerically simulating the reverse-tim e SDE for each target modality. Starting from a pure noise sample, for instance $\mathbf{x}^\textbf{L}_T \sim \mathcal{N}(\mathbf{0}, \mathbf{I})$, we iteratively denoise it by repeatedly applying a discretized update rule from \(t=T\) down to \(t=0\):
        \begin{equation}
        \mathbf{x}^\textbf{L}_{t-1} = \mathbf{x}^\textbf{L}_t + \sigma^{2t} \cdot s^\textbf{L}_\theta(\mathbf{x}^\textbf{D}_t, t) + \sigma^t \epsilon_t,
        \label{eq:reverse_update}
        \end{equation}
        where $s^\textbf{L}_\theta$ is the score estimated by our network, and \(\epsilon_t \sim \mathcal{N}(\mathbf{0},\mathbf{I})\) is a random noise term.

        \paragraph{\textbf{Diffusion Optimization Objective}} Following \Cref{fig:method}, the optimization objectives of \(s_\theta^{\textbf{L}}(\tilde{\mathbf{x}}^a_t,y,t)\) is   the same   as \Cref{eq:objective-VP-SDE}, denoted as \(\mathcal{L}_\textbf{L}\). The overall optimization objectives \(\mathcal{L}_{\text{score}}\) is as follows:
                \begin{equation}
                    \mathcal{L}_{\text{score}} = \mathcal{L}_{\textbf{L}} + \mathcal{L}_{\textbf{C}} +  \mathcal{L}_{\textbf{D}},
                    \label{eq:L_score_old}
                \end{equation} 
        following \Cref{eq:objective-VP-SDE},  $\mathcal{L}_{*}$ ($*$ denotes $\textbf{L}$, $\textbf{D}$ and $\textbf{C}$) is calculated by:  
                \begin{equation}    
                    \mathcal{L}_{*}=\mathbb{E}_{\mathbf{x}^{*},\epsilon\sim\mathcal{N}(\mathbf{0},\mathbf{I}),t\sim\mathcal{U}(0,T)}\left\|s_\theta^{{*}}(\tilde{\mathbf{x}}_t^*,t)+\frac{\epsilon}{\sqrt{\lambda(t)}}\right\|^2_2
                    \label{eq:La}
                \end{equation}

        % \subsection{Multimodal EHR Decoder}
        % \label{sec:rec}
        % Furthermore, we deploy \(\mathcal{D}_*\), where \(* \in \{\textbf{L},\textbf{C},\textbf{D}\}\) to decodes the generated data, Thus, we leverage reconstruction loss to optimize  the generated data by: 
        %         \begin{equation}
        %             \mathcal{L}_{\text{rec}}  =
        %             \|\hat{\mathbf{x}}^\textbf{L}-\mathbf{L}\|^2_2 +
        %             \|\hat{\mathbf{x}}^\textbf{D}-\mathbf{D}\|^2_2+
        %             \|\hat{\mathbf{x}}^\textbf{C}-\mathbf{C}\|^2_2.
        %             \label{L_rec}
        %         \end{equation}
        % By combining \( \mathcal{L}_{\text{score}} \) and \(\mathcal{L}_{\text{rec}}\), the objective of our multimodal diffused network is \(\mathcal{L}_{\text{total}}=\mathcal{L}_{\text{score}}+\mathcal{L}_{\text{rec}}\).
        \subsection{Multimodal EHR Decoder}
        \label{sec:rec}
        The final component of the TCDiff framework is the Multimodal EHR Decoder. Its purpose is to translate the synthesized latent representations ($\tilde{\mathbf{x}}_0^*$) from the diffusion network back into a original EHR format, composed of language, continuous, and discrete data. This decoder consists of three independent sub-networks ($\mathcal{D}_\mathbf{L}, \mathcal{D}_\mathbf{C}, \mathcal{D}_\mathbf{D}$), each tailored to a specific modality. To ensure the generated data is faithful to the original, the decoders are trained concurrently with the diffusion model via a dedicated \textbf{reconstruction loss}, $\mathcal{L}_{\text{rec}}$:
                \begin{equation}
                    \mathcal{L}_{\text{rec}}  =
                    \|\mathcal{D}_\mathbf{L}(\tilde{\mathbf{x}}_0^\mathbf{L})-\mathbf{L}\|^2_2 +
                    \|\mathcal{D}_\mathbf{D}(\tilde{\mathbf{x}}_0^\mathbf{D})-\mathbf{D}\|^2_2+
                    \|\mathcal{D}_\mathbf{C}(\tilde{\mathbf{x}}_0^\mathbf{C})-\mathbf{C}\|^2_2.
                    \label{L_rec}
                \end{equation}
        By incorporating this objective, the final training loss becomes a hybrid of score-matching and reconstruction:
                \begin{equation}
                    \mathcal{L}_{\text{total}}=\mathcal{L}_{\text{score}}+\lambda_{\text{rec}}\mathcal{L}_{\text{rec}},
                \end{equation}
        where $\lambda_{\text{rec}}$ is a balancing hyperparameter. This ensures that TCDiff not only learns the correct data distribution but also produces high-fidelity, directly usable synthetic records.

        \subsection{Generation Process of New EHR Data}
        \label{sec:gen}
        Once the TCDiff framework is trained, large quantity of synthetic EHR records are generated by simulating the reverse diffusion process. This process begins with pure noise vectors and iteratively denoises them from \(t=T\) to \(t=0\). The core of this synthesis is guided by our triplex cascaded architecture, which partitions the reverse-time trajectory for any target modality (e.g., textual, $\mathbf{x}^\textbf{L}$) into three distinct phases: \textbf{Reference Modalities Stage ($t \in[T,t_2)$):} In the initial high-noise steps, the score network primarily leverages the conditioning modalities ($\mathbf{x}^\textbf{D}, \mathbf{x}^\textbf{C}$) to establish the coarse structure and global layout of the target modality $\mathbf{x}^\textbf{L}$.
        \textbf{Cross-Modal Bridging ($t \in [t_2, t_1)$):} During these intermediate steps, the cross-modal bridging blocks actively fuse information across all conditioning modalities to align their semantic representations and enforce clinical dependencies.
        \textbf{Target Modality Stage ($t \in [t_1,0)$):} In the final, low-noise steps, the process focuses on synthesizing the fine-grained, modality-specific features of the target modality $\mathbf{x}^\textbf{L}$ with high fidelity.

        This entire cascaded generation process is performed symmetrically to synthesize all modalities. Once the final denoised latent representations ${\tilde{\mathbf{x}}^\textbf{D}_0, \tilde{\mathbf{x}}^\textbf{C}_0, \tilde{\mathbf{x}}^\textbf{L}_0}$ are obtained, they are passed to the multimodal EHR decoder, which decodes them into their original feature spaces to produce the final, high-fidelity synthetic EHR data.
%-------------------------------------------------------------------------

\section{TCM-SZ1 Datasets}
\label{sec:TCM-SZ1-Datasets}
    Despite the rapid growth of research leveraging Electronic Health Records (EHRs), the availability of datasets specifically tailored for Traditional Chinese Medicine (TCM) remains notably limited, which plays a critical role in the diagnostic and therapeutic processes of TCM. To address this gap, we introduce and publicly release the \textbf{TCM-SZ1 dataset}, a comprehensive TCM-based EHR dataset with discrete, continuous and textual modalities collected from a hospital\footnote{The name of the institution is withheld to comply with the double-blind review process and will be disclosed upon publication.}, comprising approximately 60,000 patient examination records. This study was conducted in accordance with national and international ethical guidelines, including the Measures for the Ethical Review of Biomedical Research Involving Humans (China), the Good Clinical Practice (GCP) principles, and the Declaration of Helsinki. For more details, please refer to Appendix \ref{sec:appendix:TCM-SZ1-Dataset}.
    
    With the open release of the TCM-SZ1 dataset, we aim to provide a valuable resource for researchers in the biomedical informatics community, particularly promoting advanced research in multimodal data modeling, synthetic data generation, and privacy-preserving methodologies in the historically underrepresented domain of Traditional Chinese Medicine.

%%%%%%%%%%%%%%%%%%%%%%%%%%%%%%%%%%%%%%%%%%%%%%%%%%%% Experimental Results
\section{Experimental Results}
In this section, we demonstrate extensive experimental results on three datasets. For abaltion study, please refer to Appendix \ref{sec:ex:more}.

    \paragraph{\textbf{Datasets.}}
    We evaluate our framework on three distinct de-identified EHR datasets. MIMIC-III~\cite{johnson2016mimic} is a large-scale, single-center ICU benchmark from the Beth Israel Deaconess Medical Center, renowned for its comprehensive, multimodal data that includes detailed, unstructured clinical notes. eICU~\cite{Pollard2018TheEC}, a multi-center database collected from over 200 hospitals across the United States, serves as a challenging testbed for evaluating model generalization across diverse data collection practices. Complementing these, our newly introduced TCM-SZ1 dataset enables evaluation within the unique context of Traditional Chinese Medicine.

    \paragraph{\textbf{Baselines.}}
    We compare TCDiff against a suite of state-of-the-art EHR synthesis models. These include three GAN-based methods: MedGAN \cite{choi2017generating}, CorGAN \cite{torfi2020corgan}, and EHR-M-GAN \cite{li2023generating}, as well as three advanced diffusion-based methods: LDM \cite{rombach2022high}, TabDDPM \cite{kotelnikov2023tabddpm}, and FLEXGEN-EHR \cite{he2024flexible}. Detailed descriptions of these baselines are provided in Appendix \ref{sec:ex:more}.

        \begin{figure}
            \centering
            \includegraphics[width=1\linewidth]{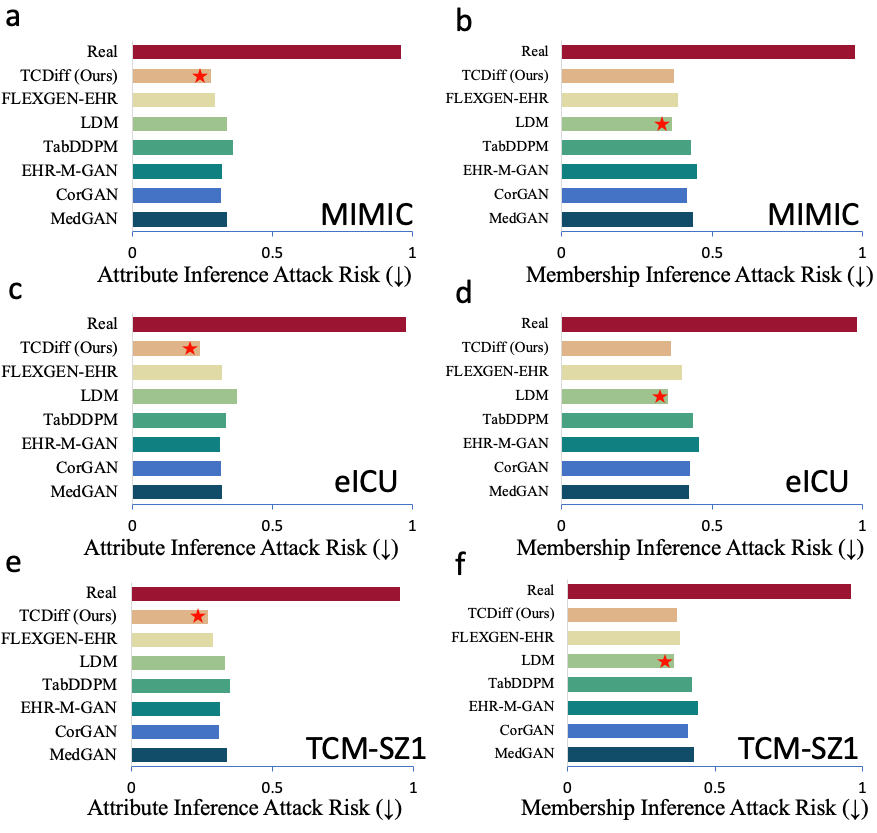}
            \caption{Privacy risk comparison. The table shows the average risk for both Attribute Inference Attacks (AIA) and Membership Inference Attacks (MIA), calculated over the 0\%-67\% missing rate spectrum. Lower is better for both metrics.}
            \label{fig:privacy}
        \end{figure}
    \paragraph{\textbf{Evaluate Metrics}} 
    To comprehensively validate the effectiveness of TCDiff, we pay attention to both fidelity and privacy. For fidelity, measuring the statistical similarity between synthetic and real EHR data, we employ three established metrics: the coefficient of determination ($R^2$, higher is better) for predictive agreement, maximum mean discrepancy (MMD, lower is better) for distributional divergence and mean squared error (MSE, lower is better) for average squared difference. As to privacy, we quantify the privacy leakage risk via attribute inference attack (AIA, lower is better) and membership inference attack (MIA, lower is better) \cite{yan2022multifaceted}.

    \paragraph{\textbf{Implementation Details}} We evaluate our model under the real-world setting where each sample has partially missing modalities. To quantify the overall degree of missingness across the dataset, we define the missing rate as: \(r_{\text{miss}} = \left(1 - \frac{\sum_{i=1}^N m_i}{N \times M} \right) \times 100\%\)
    where $N$ is the number of samples, $M = 3$ is the total number of modalities, and $m_i$ denotes the number of observed (i.e., non-missing) modalities for the $i$-th sample. In the case of three modalities, we consider eight missing rate values: \( r_{\text{miss}} \in \{0\%, 10\%, 20\%, \dots, 67\%\} \). The maximum missing rate is capped at 67\% to ensure that at least one modality is always present for each sample. Following the recent work \cite{he2024flexible}, we used k-NN imputation to pre-process the training sets for baseline models that lack inherent support for incomplete data.
    For modality-specific preprocessing, the textual modality is encoded using a pre-trained Sentence Transformer~\cite{reimers-2019-sentence-bert}, the continuous modality is normalized to zero mean and unit variance, and the categorical modality is transformed via one-hot encoding. For a fair comparison, we note that none of the existing baselines natively support textual modalities. Therefore, we apply a consistent preprocessing step across all methods by encoding textual modalities using a pre-trained Sentence Transformer \cite{reimers-2019-sentence-bert}. For MedGAN \cite{choi2017generating} only supporting discrete modalities, we use the same normalization preprocessing for the continuous modality as other baselines. More details are provided in \Cref{sec:appendix:TCM-SZ1-Dataset}.

    \subsection{Result Comparison on Data Fidelity} 
    \Cref{exp:missing_1} (for lower missing rate) and \Cref{exp:missing_2} (for higher missing rate) present a comprehensive comparison of TCDiff against state-of-the-art models on three diverse datasets under varying missing rates.
    Our methods consistently outperform the baselines in all metrics about 10\% on average.
    The results lead to the following key findings: 
        \textbf{(1) Superiority of Diffusion-based Models in Handling Heterogeneity.}
        Across all datasets, a clear trend emerges: diffusion-based models (TabDDPM, LDM, and our TCDiff) consistently form the top tier, significantly surpassing earlier GAN-based methods. This suggests that the diffusion paradigm's inherent capacity for modeling complex high-dimensional distributions provides a more powerful foundation for capturing the intrinsic characteristics of heterogeneous EHR data.
        \textbf{(2) State-of-the-Art Performance Driven by Cascaded Architecture.}
        While diffusion models as a class perform well, TCDiff outperforms even other diffusion baselines (LDM and TabDDPM) across all metrics and scenarios. The results strongly indicate that the Cross-Modal Bridging stage in our architecture is highly effective at capturing the intricate cross-modal dependencies that other models fail to handle, leading to a more clinically coherent and realistic data synthesis.
        \textbf{(3) Strong Robustness to Missing Modalities.}
        The most compelling evidence of TCDiff's advantage is revealed under high data incompleteness (\Cref{exp:missing_2}). As the missing rate increases from 0\% to an extreme 67\%, the performance of baseline models degrades sharply. In contrast, TCDiff exhibits a remarkably stable performance curve with minimal degradation. This validates the robustness of our multi-stage generative design, which is architecturally flexible to incomplete data. The results confirm that our framework can maintain high-fidelity generation even when a substantial portion of the source data is unavailable, directly addressing the critical limitation in real-world scenarios.

    \subsection{Result Comparision on Data Privacy}
    \Cref{fig:privacy} summarizes the privacy assessment, reporting the average risk scores \cite{yan2022multifaceted} for both Attribute Inference Attacks (AIA) and Membership Inference Attacks (MIA) across the 0\%-67\% missing rate spectrum.
    The results reveal a key insight. Across both AIA and MIA metrics, our TCDiff and the LDM baseline consistently demonstrate the lowest risk scores. LDM's strong privacy performance is likely a byproduct of its lower data fidelity. The generated samples by it are less faithful to the real patient records, making them inherently harder for an adversary to exploit in either attack scenario.
    In contrast, TCDiff provides robust protection against both types of privacy attacks while simultaneously delivering state-of-the-art data fidelity (as shown in \Cref{exp:missing_1,exp:missing_2}). This highlights TCDiff's primary advantage: it successfully navigates the critical privacy-utility trade-off, providing comprehensive privacy guarantees without compromising the data quality.

    \section{Conclusion}
    In this paper, we presented TCDiff, a novel triplex cascaded diffusion framework designed to overcome critical bottlenecks in high-fidelity EHR synthesis. Its multi-stage architecture systematically addresses the key challenges of modeling heterogeneous modalities (including clinical text), capturing complex inter-modal dependencies, and robustly handling data incompleteness. 
    Our extensive validation was performed on both public benchmarks and the newly introduced TCM-SZ1, a large-scale dataset we constructed to spur research in this area. The results confirm TCDiff's superiority, demonstrating state-of-the-art data fidelity across a wide spectrum of missing rates and strong privacy preservation. This underscores its effectiveness, robustness, and generalizability as a practical solution for real-world healthcare scenarios.

\begin{acks}
    This paper was supported by National Key Research and Development Program of China (2023YFC3502902, 2022YFB2703300, 2022YF
    C2503903), National Natural Science Foundation of China under Grants 62436006. 
\end{acks}

\bibliographystyle{ACM-Reference-Format}
\bibliography{sample-base}

\appendix

\section{Details of TCM-SZ1 Dataset}
\label{sec:appendix:TCM-SZ1-Dataset}
\paragraph{\textbf{Details}}The TCM-SZ1 dataset includes multimodal EHR data consisting of continuous, discrete, and textual modalities, providing a rich source of information for diverse healthcare analyses:
    
    \begin{itemize}

        \item \textbf{Discrete modality}: The discrete data modality encompasses a range of categorical variables describing patients' states and characteristics, including:
        \textit{Nutrition}, \textit{Body Form}, \textit{Facial Expression}, \textit{Emotion}, \textit{Posture}, \textit{Spirit}, \textit{text}, \textit{Cooperation during Examination}, \textit{Communication}, \textit{Tongue Condition}, \textit{Tongue Fur}, and \textit{Veins}. Each discrete attribute is represented using a one-hot encoding scheme and concatenated to form the final discrete feature representation. For instance, the attribute ``\textit{Nutrition}'' contains categories such as ``\textit{Malnutrition}'', ``\textit{Moderate Nutrition}'', ``\textit{Overnutrition}'', and ``\textit{Good Nutrition}''.
        % 这里怎么渲染中文？   

        \item \textbf{Continuous modality}: This modality includes vital signs and physiological measurements such as body temperature, heart rate, respiratory rate, systolic blood pressure, and diastolic blood pressure. Each continuous feature is normalized within the range [0, 1] to facilitate downstream analysis and modeling. For example, a record might contain normalized values like Temperature (0.37), Heart Rate (0.80), Respiratory Rate (0.18), Systolic Blood Pressure (0.55), and Diastolic Blood Pressure (0.36).

        \item \textbf{Textual modality}: Textual information in TCM provides essential semantic contexts, including patients' chief complaints and corresponding TCM diagnoses. For instance, a typical textual input includes a chief complaint such as ``\textit{Intermittent hematochezia for 3 years, colon polyps identified 4 days ago.}'' and a corresponding TCM diagnosis ``\textit{Intestinal polyps, damp-heat in intestines}''.
        % \item  In our dataset, textual descriptions are encoded into fixed-dimensional embedding vectors using the sentence-transformer model~\cite{reimers-2019-sentence-bert}.
    \end{itemize} 

    As a concrete example, a complete record from the TCM-SZ1 dataset could be represented as follows:
    
    %%描述一下时序
    \begin{itemize}
        
        \item \textbf{Discrete}: Nutrition (\textit{``Good Nutritio''}), BodyForm (\textit{``Moderate Body Form''}), FacialExpression (\textit{``Normal''}), Emotion (\textit{``Calm''}), Posture (\textit{``Active''}), Spirit (\textit{``Energetic''}), text (\textit{``Clear and Fluent''}), Cooperation (\textit{``Cooperative''}), Communication (\textit{``Relevant Answers''}), Tongue (\textit{``Light Red Tongue with Teeth Marks''}), Fur (\textit{``Thin White Coating''}), Veins (\textit{``Fine Pulse''}).
        
        \item \textbf{Continuous}: Temperature = 37.2°C, Heart Rate = 80 bpm, Respiratory Rate = 18 bpm, Systolic Blood Pressure = 111 mmHg, Diastolic Blood Pressure = 73 mmHg.

        \item \textbf{Textual}: Chief Complaint (\textit{``Pain in waist and hip for over 4 months''}); TCM Diagnosis (\textit{``Lumbago, Qi stagnation and blood stasis''}).
    \end{itemize}

\begin{table}[t]
    \centering
    \small
    \caption{Ablation study about our architeture on fidelity, calculated over the 0\%-67\% missing rate spectrum. 
    Each cell reports average scores: $\mathbf{R^2}$ $(\uparrow)$ / MMD $(\downarrow)$ / MSE $(\downarrow)$. 
    \textbf{Bold} indicates the best result in each group.}
    \label{tab:ablation-fid}
    \resizebox{\linewidth}{!}{
    \begin{tabular}{llc}
        \toprule
        \textbf{Dataset} & \textbf{Method} & \textbf{Fidelity} \\
        \midrule
        \multirow{4}{*}{MIMIC-III~\cite{johnson2016mimic}}  
            & (iv) \textbf{Ours}         & \textbf{0.723 / 0.569 / 0.733}  \\
            & (iii)  w/o Triplex Design  & 0.695 / 0.599 / 0.785 \\
            & (ii)   w/o Cascade Design  & 0.652 / 0.642 / 0.812 \\
            & (i)    Base model          & 0.608 / 0.686 / 0.845 \\
        \midrule
        \multirow{4}{*}{eICU~\cite{Pollard2018TheEC}}  
            & (iv) \textbf{Ours}         & \textbf{0.717 / 0.589 / 0.745}  \\
            & (iii)  w/o Triplex Design  & 0.693 / 0.625 / 0.801 \\
            & (ii)   w/o Cascade Design  & 0.641 / 0.657 / 0.831 \\ 
            & (i)    Base model          & 0.617 / 0.682 / 0.863 \\
        \midrule
        \multirow{4}{*}{TCM-SZ1}  
            & (iv) \textbf{Ours}         & \textbf{0.728 / 0.563 / 0.713} \\
            & (iii)  w/o Triplex Design  & 0.702 / 0.625 / 0.813 \\
            & (ii)   w/o Cascade Design  & 0.683 / 0.657 / 0.841 \\
            & (i)    Base model          & 0.647 / 0.692 / 0.872 \\
        \bottomrule
    \end{tabular}
    }
\end{table}

\begin{table}[t]
\centering
\small
\caption{Ablation study about our architeture on privacy, calculated over the 0\%-67\% missing rate spectrum. 
Each cell reports average scores: Attribute / Membership inference risks $(\downarrow)$. 
\textbf{Bold} indicates the best result in each group.}
\label{tab:ablation-pri}
\resizebox{\linewidth}{!}{
\begin{tabular}{llc}
    \toprule
    \textbf{Dataset} & \textbf{Method} & \textbf{Privacy} \\
    \midrule
    \multirow{4}{*}{MIMIC-III~\cite{johnson2016mimic}}  
        & (iv) \textbf{Ours}         & \textbf{0.2} / 0.374 \\
        & (iii)  w/o Triplex Design \quad \quad \quad & 0.292 / 0.364  \\
        & (ii)   w/o Cascade Design \quad \quad \quad & 0.293 / 0.359 \\
        & (i)    Base model         \quad \quad \quad & 0.315 / \textbf{0.358} \\
    \midrule
    \multirow{4}{*}{eICU~\cite{Pollard2018TheEC}}  
        & (iv) \textbf{Ours}         & \textbf{0.265} / 0.368 \\
        & (iii)  w/o Triplex Design \quad \quad \quad & 0.278 / 0.362  \\
        & (ii)   w/o Cascade Design \quad \quad \quad & 0.284 / 0.359 \\
        & (i)    Base model         \quad \quad \quad & 0.309 / \textbf{0.357} \\
    \midrule
    \multirow{4}{*}{TCM-SZ1}  
        & (iv) \textbf{Ours}         & \textbf{0.242} / 0.362  \\
        & (iii)  w/o Triplex Design \quad \quad \quad & 0.288  / \textbf{0.357}  \\
        & (ii)   w/o Cascade Design \quad \quad \quad & 0.290 /  0.360 \\
        & (i)    Base model         \quad \quad \quad & 0.311 /  \textbf{0.357} \\
    \bottomrule
\end{tabular}
}
\end{table}

\section{More Experimental Results}
\label{sec:ex:more}
\subsection{More Details of Experiments}
All experiments are conducted on an NVIDIA RTX 4090 GPU (24\,GB). For diffusion-based models, we adopt 100 discrete time steps. In the cascaded diffusion architecture, the transition points are hyperparameters set to \( t_1 = 20 \) and \( t_2 = 80 \). The noise scaling coefficient in the stochastic differential equation (SDE) is fixed at \( \sigma = 25 \). All models are trained for a maximum of 300 epochs using the Adam optimizer, with a learning rate of \(1 \times 10^{-4}\). We apply early stopping with a patience of 20 epochs based on the validation loss. 
\subsection{Ablation Study}
\paragraph{\textbf{Impact of Different Configurations}}To validate the effectiveness of different architectural configurations in our diffusion-based model, we conduct a comprehensive ablation study with four comparative variants:
\textbf{(i)} \textbf{Base Model}: A single-layer score-based diffusion model following the standard architecture \cite{song2020score}, which directly predicts the target modality without any cascaded design.
\textbf{(ii)} \textbf{w/o Cascade Design}: A single-layer model equipped with the Triplex structure but without the cascaded diffusion mechanism, thus limiting the inter-modal information flow across stages.
\textbf{(iii)} \textbf{w/o Triplex Design}: A model that incorporates the cascaded diffusion mechanism but replaces the Triplex design with a simpler concatenation-based fusion, weakening the expressive capacity of modality interactions.
\textbf{(iv)} \textbf{Ours}: The full version of our model, which integrates both the cascaded diffusion mechanism and the Triplex design for enhanced multimodal learning and privacy preservation.

As shown in \Cref{tab:ablation-fid,tab:ablation-pri}, our full model (iv) consistently achieves the best performance across both privacy and fidelity metrics on the MIMIC-III and TCM-SZ1 datasets. Notably, while removing certain modules (e.g., Triplex or cascade design) may lead to slightly improved privacy scores in isolated cases, such configurations exhibit a significant drop in data fidelity. This indicates that these ablated models fail to effectively balance utility and privacy, reinforcing the importance of integrating both architectural components for high-quality synthetic data generation.

\paragraph{\textbf{Effects of the Multimodal EHR Encoder and Decoder}} 
We investigate the individual contributions of the multimodal EHR encoder (\(\mathcal{E}_*\)) and the multimodal EHR decoder (\(\mathcal{D}_*\)), where \(*\) denotes the discrete, continuous, and textual modalities. We perform an ablation study using four model variants. All configurations maintain the full triplex cascaded diffusion backbone, differing only in whether the encoder and/or decoder modules are included.

As shown in \Cref{tab:ablation-encdec}, incorporating both \(\mathcal{E}_*\) and \(\mathcal{D}_*\) yields the best overall performance on both MIMIC-III and TCM-SZ1 datasets, demonstrating the importance of these modules for effective multimodal representation learning and high-fidelity generation. Specifically, removing either the encoder or the decoder leads to a noticeable degradation in $R^2$ scores and an increase in both MMD and MSE, indicating diminished reconstruction quality and weakened modality alignment.

Interestingly, omitting the encoder (\ding{55} for \(\mathcal{E}_*\)) while retaining the decoder (\ding{51} for \(\mathcal{D}_*\)) results in better performance than the reverse configuration. This suggests that the decoder plays a more critical role in preserving the structural integrity of the generated EHR data, while the encoder mainly contributes to cross-modal feature integration and latent consistency.

In the worst-case setting, where both modules are removed (\ding{55}, \ding{55}), the model suffers from the largest drop in $R^2$ and the highest MMD and MSE values. This confirms that these modules are not merely auxiliary components, but essential for achieving high-quality synthetic data generation.

Overall, this ablation study highlights the complementary roles of the multimodal feature extractor and decoder in enhancing generation fidelity and supports their inclusion in the full model design.

\paragraph{\textbf{Impact of the training strategy}}
To validate the effectiveness of our proposed training strategy, we conducted an ablation study under 50\% missing rate, with results shown in \Cref{tab:ablation-training}. (1) Comparing (B) with (A), we observe that using a standard k-NN imputer is significantly better than naive zero-filling. (2) More importantly, our full model (D) outperforms the static k-NN approach (B) by a large margin, demonstrating the superiority of the iterative refinement process. (3) Finally, comparing (D) with (C), the performance drop of the ``cold start'' version highlights the necessity of the warm-up phase for stabilizing the training. These results collectively confirm that each component of our self-consistent training strategy is essential for achieving the best performance.

\begin{table}[t]
    \centering
    \small
    \caption{Ablation study on the training strategy under 50\% missing rate.
    Each cell reports average scores: $R^2$ $(\uparrow)$ / MMD $(\downarrow)$ / MSE $(\downarrow)$. 
    \textbf{Bold} indicates the best result in each group.}
    \label{tab:ablation-training}
    \resizebox{\linewidth}{!}{
    \begin{tabular}{llc}
        \toprule
        \textbf{Dataset} & \textbf{Method} & \textbf{Fidelity} \\
        \midrule
        \multirow{4}{*}{MIMIC-III~\cite{johnson2016mimic}}  
        & (i) TCDiff w/ Zero Imputation   & 0.634 / 0.819 / 0.963 \\
        & (ii)TCDiff w/ k-NN Imputation   & 0.681 / 0.705 / 0.854 \\
        & (iii)  TCDiff w/o Warm-up       & 0.715 / 0.621 / 0.792 \\
        & (iv)   TCDiff (Full)            & \textbf{0.720 / 0.613 / 0.776} \\
        \midrule
        \multirow{4}{*}{eICU~\cite{Pollard2018TheEC}}  
        & (i) TCDiff w/ Zero Imputation   & 0.615 / 0.841 / 0.997 \\
        & (ii)TCDiff w/ k-NN Imputation   & 0.669 / 0.728 / 0.882 \\
        & (iii)  TCDiff w/o Warm-up       & 0.708 / 0.645 / 0.801 \\
        & (iv)   TCDiff (Full)            & \textbf{0.714 / 0.633 / 0.788} \\
        \midrule
        \multirow{4}{*}{TCM-SZ1}  
        & (i) TCDiff w/ Zero Imputation   & 0.592 / 0.906 / 1.125 \\
        & (ii)TCDiff w/ k-NN Imputation   & 0.658 / 0.753 / 0.974 \\
        & (iii)  TCDiff w/o Warm-up       & 0.696 / 0.651 / 0.889 \\
        & (iv)   TCDiff (Full)            & \textbf{0.704 / 0.633 / 0.856} \\
        \bottomrule
    \end{tabular}
    }
\end{table}

\begin{table}[t]
\centering
\small
\caption{Ablation study on the effects of the multimodal EHR encoder 
and multimodal EHR decoder. We compare performance calculated over the 0\%-67\% missing rate spectrum. Each cell reports average scores:  $\mathbf{R^2}$ $(\uparrow)$ / MMD $(\downarrow)$ / MSE $(\downarrow)$. \textbf{Bold} indicates the best result in each group.}
\resizebox{\linewidth}{!}{
\begin{tabular}{c|cc|c}
    \toprule
        Datasets & \(\mathcal{E}_*\) & \(\mathcal{D}_*\) & Fidelity \\ 
    \midrule
    \midrule
    \multirow{4}{*}{MIMIC-III~\cite{johnson2016mimic}} 
        & \quad \ding{51} \quad & \quad \ding{51} \quad & \quad \quad  \textbf{0.723 / 0.569 / 0.733} \quad \quad\\  
        & \quad \ding{55} \quad & \quad \ding{51} \quad & \quad \quad 0.715 / 0.602 / 0.785           \quad \quad\\
        & \quad \ding{51} \quad & \quad \ding{55} \quad & \quad \quad 0.672 / 0.637 / 0.812           \quad \quad\\ 
        & \quad \ding{55} \quad & \quad \ding{55} \quad & \quad \quad 0.648 / 0.666 / 0.831           \quad \quad\\ 
    \midrule
    \midrule
    \multirow{4}{*}{eICU~\cite{Pollard2018TheEC}}  
        & \quad \ding{51} \quad & \quad \ding{51} \quad & \quad \quad \textbf{0.717 / 0.589 / 0.745}  \quad \quad\\
        & \quad \ding{55} \quad & \quad \ding{51} \quad & \quad \quad 0.703 / 0.625 / 0.801 \quad \quad\\
        & \quad \ding{51} \quad & \quad \ding{55} \quad & \quad \quad 0.661 / 0.657 / 0.831 \quad \quad\\ 
        & \quad \ding{55} \quad & \quad \ding{55} \quad & \quad \quad 0.637 / 0.682 / 0.863 \quad \quad\\
    \midrule
    \midrule
    \multirow{4}{*}{TCM-SZ1} 
        & \quad \ding{51} \quad & \quad \ding{51} \quad & \quad \quad \textbf{0.728 / 0.563 / 0.713} \quad \quad\\  
        & \quad \ding{55} \quad & \quad \ding{51} \quad & \quad \quad 0.728 / 0.625 / 0.813 \quad \quad\\
        & \quad \ding{51} \quad & \quad \ding{55} \quad & \quad \quad 0.684 / 0.657 / 0.841 \quad \quad\\ 
        & \quad \ding{55} \quad & \quad \ding{55} \quad & \quad \quad 0.661 / 0.692 / 0.872 \quad \quad\\
    \bottomrule
\end{tabular}
}
\label{tab:ablation-encdec}
\end{table}

The descriptions of the baselines are as follows:
\begin{itemize}
        
    \item \textbf{MedGAN}~\cite{choi2017generating}: A GAN-based framework specifically tailored for synthesizing medical records. It generates low-dimensional discrete synthetic data, which is subsequently decoded into original feature spaces via an autoencoder component. It has been widely adopted for generating realistic yet synthetic patient records.
    
    \item \textbf{CorGAN}~\cite{torfi2020corgan}: Another GAN-based model that integrates Convolutional Generative Adversarial Networks (ConvGANs) and Convolutional Autoencoders. It emphasizes synthesizing high-quality medical data by leveraging convolutional architectures, enabling effective capture of local dependencies within medical records.
    
    \item \textbf{EHR-M-GAN}~\cite{li2023generating}: A GAN-based model explicitly designed for longitudinal and heterogeneous EHR data. This framework effectively handles multiple data types(continuous and discrete), thereby generating realistic and temporally consistent patient trajectories.
    
    \item \textbf{TabDDPM}~\cite{kotelnikov2023tabddpm}: A diffusion probabilistic model specifically developed for tabular data synthesis. TabDDPM employs a diffusion-based generative process suitable for handling heterogeneous feature types(continuous and discrete), effectively overcoming several challenges inherent in traditional GAN-based approaches, such as mode collapse and unstable training.
    
    \item \textbf{LDM}~\cite{rombach2022high}: Latent Diffusion Model decomposes the generative process into two sequential steps: first encoding high-dimensional data into a latent space with an autoencoder, and then applying diffusion models within this latent space. This approach greatly reduces the computational complexity while maintaining high-fidelity synthetic data generation.

    \item \textbf{FLEXGEN-EHR}~\cite{he2024flexible}: A latent diffusion model tailored for heterogeneous tabular EHRs(continuous and discrete), equipped with the capability of handling missing modalities in an integrative learning framework. This work defines an optimal transport module to align and accentuate the common feature space of heterogeneity of EHRs.
    
\end{itemize}

\section{Discussion}
\subsection{Real-world Applications}

The TCDiff framework demonstrates strong potential for the deployment in diverse clinical and biomedical research contexts. By addressing the challenge of incomplete data, it enhances the reliability of electronic health record (EHR) systems, particularly in scenarios where routine clinical documentation is fragmented. The ability to reconstruct missing modalities—such as omitted lab results or incomplete symptom narratives—can significantly improve the continuity of patient records. Beyond imputation, TCDiff serves as a powerful tool for generating privacy-preserving synthetic data, offering a practical solution for secure data sharing under stringent regulatory constraints (e.g., HIPAA, GDPR). This is especially valuable in settings where data access is limited due to privacy concerns or institutional policies, such as Traditional Chinese Medicine (TCM) hospitals. By retaining key statistical and clinical characteristics, the synthetic data produced by TCDiff can support research reproducibility, benchmarking, and model pretraining without compromising patient confidentiality. In addition, the model's cross-modal generation capability opens doors for innovative applications in clinical reasoning and medical text understanding. The ability to infer structured variables from text—or vice versa—facilitates semantic alignment across modalities, supporting tasks such as automatic coding, clinical summarization, and intelligent triage in semi-structured EHR systems.

\subsection{Limitations and Future Work}
While TCDiff demonstrates promising performance, several limitations warrant further investigation to advance its practical deployment. First, our current framework inherently assumes fixed-length tabular inputs, which constrains its applicability to longitudinal or event-based EHRs characterized by irregular temporal intervals and variable-length sequences. Enhancing the model's temporal representational capacity, for instance, by enabling each modality to have its own cross-modal bridging moment, dynamically balancing the importance of different modalities during the generation process, could serve as a meaningful extension to capture complex clinical trajectories and evolving patient states.
Second, although we have validated TCDiff on both public (MIMIC-III and eICU) and proprietary (TCM-SZ1) datasets, its generalizability across diverse healthcare ecosystems remains unproven. Systematic evaluation on non-hospital data sources, including continuous monitoring streams from wearable devices, home-based telehealth systems, and insurance claims databases, would better test its robustness against heterogeneous data quality levels, varying sampling frequencies, and domain-specific modality characteristics. Such cross-domain validation is crucial for assessing the framework's adaptability to real-world healthcare data landscapes.
Most critically, the fundamental challenge of balancing task-specific utility requirements with EHRs' stringent privacy constraints remains unresolved. The development of a comprehensive evaluation framework that effectively addresses multimodal heterogeneity, enables cross-scenario generalizability, and achieves optimal utility-privacy trade-offs remains a critical research frontier in healthcare analytics.

\end{document}